\documentclass[lettersize,journal]{IEEEtran}
\usepackage{amsmath,amsfonts}
\usepackage{algorithmic}
\usepackage{array}
\usepackage[caption=false,font=normalsize,labelfont=sf,textfont=sf]{subfig}
\usepackage{textcomp}
\usepackage{stfloats}
\usepackage{url}
\usepackage{verbatim}
\usepackage{graphicx}
\usepackage{cite}
\usepackage{url}            
\usepackage{booktabs}       
\usepackage{amssymb}
\usepackage{enumitem}
\usepackage{verbatim}
\usepackage{multirow}
\usepackage{marvosym}
\usepackage{makecell}
\usepackage{bm}
\usepackage{wrapfig}
\usepackage{colortbl}

\usepackage{xcolor}         
\usepackage{pifont}
\usepackage[ruled, lined, linesnumbered, commentsnumbered, longend]{algorithm2e}
\usepackage{amsmath}
\newcommand{\argmax}{\operatornamewithlimits{argmax}}
\newcommand{\argmin}{\operatornamewithlimits{argmin}}
\newcommand{\PreserveBackslash}[1]{\let\temp=\\#1\let\\=\temp}
\newcolumntype{C}[1]{>{\PreserveBackslash\centering}p{#1}}
\newcolumntype{R}[1]{>{\PreserveBackslash\raggedleft}p{#1}}
\newcolumntype{L}[1]{>{\PreserveBackslash\raggedright}p{#1}}

\begin{document}

\title{Few-Shot Anomaly Detection via Category-Agnostic Registration Learning}

\author{Chaoqin Huang,
Haoyan Guan,
Aofan Jiang,
Ya Zhang,
Michael Spratling,
Xinchao Wang, 
Yanfeng Wang

\IEEEcompsocitemizethanks{
\IEEEcompsocthanksitem This work was supported in part by the National Key Research and Development Program of China under Grant 2022ZD0160702; in part by the STCSM under Grant 22DZ2229005; in part by the 111 Plan under Grant BP0719010; in part by the State Key Laboratory of UHD Video and Audio Production and Presentation; in part by the Ministry of Education Singapore, under its Academic Research Fund Tier 2, under Award MOE-T2EP20122-0006; and in part by the National Research Foundation, Singapore, under its AI Singapore Program (AISG), under Award AISG2-RP-2021023.
(Chaoqin Huang and Haoyan Guan contributed equally to this work.) (Corresponding authors: Yanfeng Wang; Ya Zhang.)
\IEEEcompsocthanksitem Chaoqin Huang is with the College of Information Communications, National University of Defense Technology, Wuhan, China. This work was done while he was with the Cooperative Medianet Innovation Center, Shanghai Jiao Tong University, Shanghai, China, and with the National University of Singapore, Singapore. E-mail: huangchaoqin@nudt.edu.cn.
\IEEEcompsocthanksitem Haoyan Guan and Michael Spratling are with the Department of Informatics, King's College London, London, UK. E-mail: \{haoyan.guan, michael.spratling\}@kcl.ac.uk.
\IEEEcompsocthanksitem Aofan Jiang, Ya Zhang and Yanfeng Wang are with the Cooperative Medianet Innovation Center, Shanghai Jiao Tong University, Shanghai, China, also with Shanghai Artificial Intelligence Laboratory, Shanghai, China. E-mail: \{stillunnamed, ya\_zhang, wangyanfeng622\}@sjtu.edu.cn.
\IEEEcompsocthanksitem Xinchao Wang is with the National University of Singapore, Singapore. E-mail: xinchao@nus.edu.sg.
}}

\markboth{IEEE Transactions on Neural Networks and Learning Systems}%
{Huang \MakeLowercase{\textit{et al.}}: Few-Shot Anomaly Detection via Category-Agnostic Registration Learning}


\maketitle

\begin{abstract}
Most existing anomaly detection (AD) methods require a dedicated model for each category. Such a paradigm, despite its promising results, is computationally expensive and inefficient, thereby failing to meet the requirements for realworld applications. Inspired by how humans detect anomalies, by comparing a query image to known normal ones, this article proposes a novel few-shot AD (FSAD) framework. Using a training set of normal images from various categories, registration, aiming to align normal images of the same categories, is leveraged as the proxy task for self-supervised category-agnostic representation learning. At test time, an image and its corresponding support set, consisting of a few normal images from the same category, are supplied, and anomalies are identified by comparing the registered features of the test image to its corresponding support image features. Such a setup enables the model to generalize to novel test categories. It is, to our best knowledge, the first FSAD method that requires no model fine-tuning for novel categories: enabling a single model to be applied to all categories. Extensive experiments demonstrate the effectiveness of the proposed method. Particularly, it improves the current state-of-the-art (SOTA) for FSAD by 11.3\% and 8.3\% on the MVTec and MPDD benchmarks, respectively. The source code is available at \url{https://github.com/Haoyan-Guan/CAReg}.
\end{abstract}



\section{Introduction}
\label{intro}
\IEEEPARstart{A}{nomaly} detection (AD) has recently drawn increasing attention due to its wide range of applications in defect detection~\cite{matsubara2018anomaly}, autonomous driving~\cite{eykholt2018robust,tian2022pixel}, and medical diagnosis~\cite{zhang2020viral,su2021few}. Recent studies have mainly pursued an unsupervised learning paradigm that learns with normal samples only, due to the challenge to collect an exhaustive set of anomalous samples. A typical solution is to model the feature distribution of the normal samples, and images that do not conform to this distribution are considered anomalous~\cite{zong2018deep,gong2019memorizing,metaformer,golan2018deep,ARNet,MKD,focus,scholkopf2001estimating,yi2020patch}.

\begin{figure*}[t]
\centering
\includegraphics[width=0.98\textwidth]{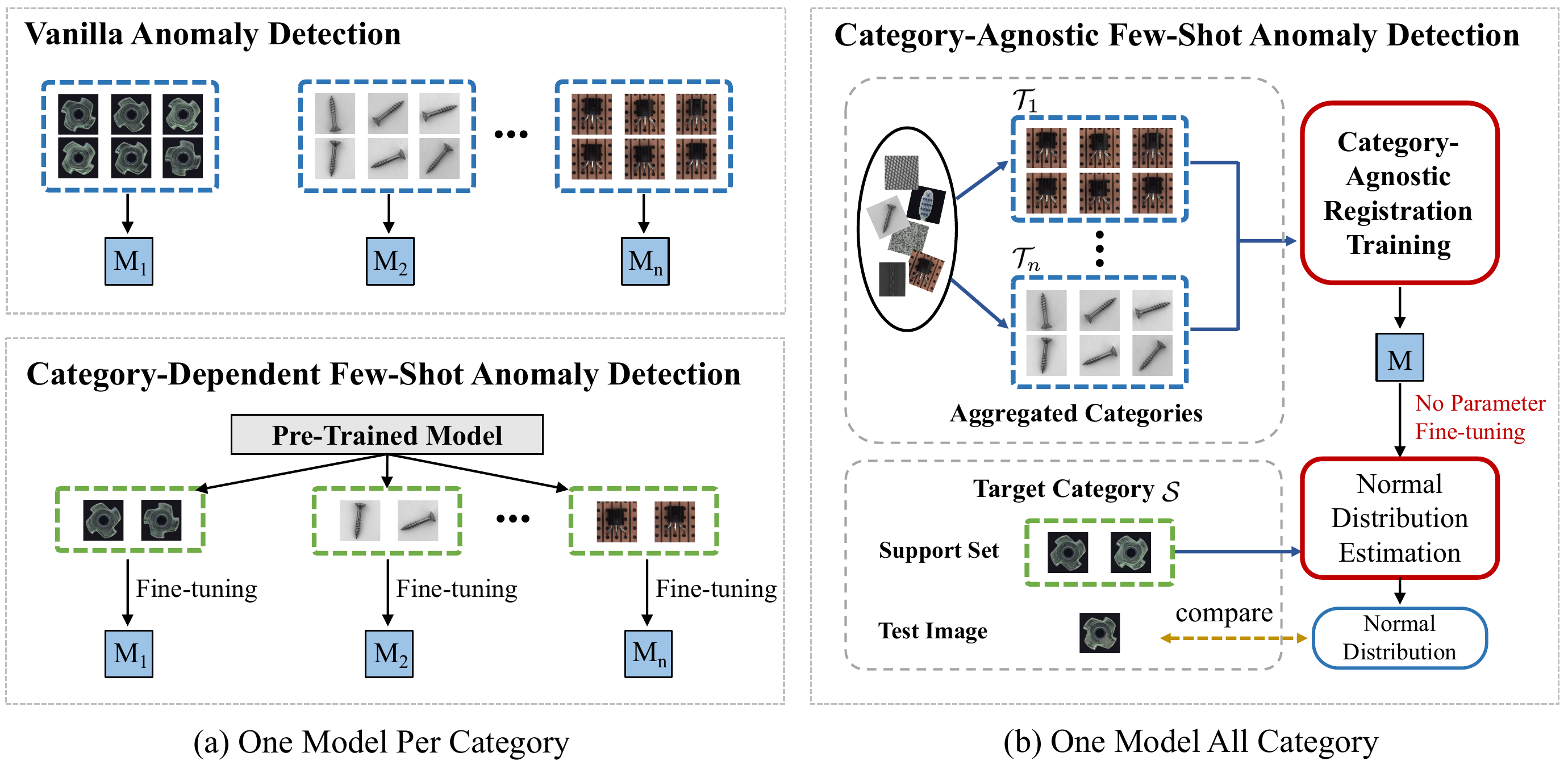}
\caption{(a) One-model-per-category paradigm for the vanilla AD and FSAD. (b) One-model-all-category paradigm for the proposed category-agnostic FSAD.}
\label{img:intro}
\end{figure*}

Most existing AD methods follow a `one-model-per-category' paradigm that trains an individual model for each category, as depicted in Fig.~\ref{img:intro} (a). Such a process requires hundreds or thousands of images which are often prohibitive to obtain in real world scenarios. To reduce this burden of data collection, few-shot anomaly detection (FSAD) has been proposed, where only a few normal images are available for each category during the fine-tuning of a pre-trained model~\cite{TDG,DiffNet}. Early attempts include either leveraging various transformations to augment the few-shot data \cite{TDG} or introducing a lighter model to estimate the normal distribution \cite{DiffNet} so as to avoid overfitting. However, the above approaches still follow the one-model-per-category learning paradigm.

Inspired by how human beings detect anomalies, this paper explores a one-model-all-category paradigm for FSAD, \emph{i.e.,} training a category-agnostic AD model. To find the anomalies in an image, humans simply compare it to a normal exemplar and identify any difference. This approach requires no knowledge of the image category because \emph{comparison} is naturally category agnostic. We thus model AD as a \emph{comparison} task to make it category agnostic. To deal with the fact that objects may appear in different positions, orientations, or poses, the images are first transformed into one coordinate system through a form of \emph{registration}~\cite{brown1992survey,Barbara2003Image,peng2011brainaligner} to facilitate the comparison. 

An overview of the proposed \underline{C}ategory-\underline{A}gnostic \underline{Reg}istration (\texttt{CAReg}) framework for FSAD is shown in Fig.~\ref{img:intro} (b). A Siamese network~\cite{chen2021exploring} containing three spatial transformer network blocks~\cite{STN} is employed for registration. For better robustness, we perform registration at the feature level, instead of pixel-by-pixel image registration~\cite{peng2011brainaligner}. The feature-level registration loss, a relaxed version of the image-level pixel-wise registration loss, is used to minimize the cosine distance of the features from images of the same category. The trained model has the ability to perform registration across different categories and is able to be used at test-time directly on novel categories. Anomalies are detected by comparing the registered features of the novel test category image and its corresponding support set (a few normal images from the target category). The comparison process first estimates the normal distribution based on the support set, which can be implemented with statistical-based normal distribution estimators~\cite{defard2021padim}, and then fits the test samples to the normal distribution. Samples that fall out of the normal distribution are considered anomalies.

We validate the effectiveness of the proposed \texttt{CAReg} framework on two challenging benchmark datasets for industrial defect detection, MVTec~\cite{bergmann2019mvtec} and MPDD~\cite{jezek2021deep}. \texttt{CAReg} is shown to outperform the state-of-the-art FSAD methods~\cite{TDG,DiffNet} in terms of AUC on 2-shot, 4-shot, and 8-shot scenarios, respectively, by 9.6\%, 12.4\% and 11.9\% on MVTec and 7.1\%, 9.1\% and 8.8\% on MPDD. We also experiment with three SOTA normal distribution estimation methods: PaDim~\cite{defard2021padim}, OrthoAD~\cite{orthoad}, and PatchCore~\cite{patchcore}. Our experimental results show that \texttt{CAReg} is able to improve the average AUC for all three normal distribution estimation methods, by 10.9\%, 15.4\% and 8.5\% on MVTec, and by 12.1\%, 4.4\% and 8.0\% on MPDD, for PaDim, OrthoAD and PatchCore, respectively. 

Below we summarize the main contributions:
\begin{itemize} 
    \item \texttt{CAReg} is the first one-model-all-category framework for FSAD, requiring no abnormal images at training and no parameter fine-tuning at testing.
    \item Feature registration is leveraged as the proxy task to train AD models, which demonstrate increased generalizability across different categories and datasets.
    \item \texttt{CAReg} outperforms the state-of-the-art FSAD methods on both AD and anomaly localization tasks with standard benchmark datasets.
\end{itemize}

Compared to our previous method, RegAD, presented at ECCV~2022~\cite{regad} as an oral presentation, the proposed approach has the following technical improvements:
\begin{itemize}
\item \texttt{CAReg} is designed as a framework for FSAD, which can flexibly adopt various normal distribution estimators in addition to PaDim that was used in our previous work. 
\item \texttt{CAReg} introduces a modified feature registration module to improve the model's robustness. Instead of registration referenced by only one single image, accumulated features from each category are used as registration references.
\item A Wasserstein distance-based method is proposed to flexibly, and automatically, choose the most suitable data augmentations during the normal distribution estimation stage, improving the accommodation of different distribution characteristics among categories in the FSAD scenario.
\item Results of more experiments are reported both for evaluating performance, and assessing the method through ablation studies.
\end{itemize}

The rest of the paper is organized as follows. Sec.~\ref{sec:related work} reviews existing works related to AD and few-shot learning. Sec.~\ref{sec:formulation} formulates the problem of FSAD. Sec.~\ref{sec:pretrain} describes the category-agnostic registration training used for \texttt{CAReg}. Sec.~\ref{sec:ad} describes an AD system combining \texttt{CAReg} with multiple normal distribution estimators.  Sec.~\ref{sec:experiment} presents the experimental results on several real-world datasets, showing the effectiveness of our method in both FSAD and anomaly localization. Finally, conclusions are drawn in Sec.~\ref{sec:conclusion}.
\section{Related Work} \label{sec:related work}
\subsection{Anomaly Detection} 
AD~\cite{pang2022editorial} has a wide range of applications, encompassing still image AD~\cite{golan2018deep,ARNet,gong2019memorizing,metaformer} and anomalous human behavior detection~\cite{fang2020anomaly,ramachandra2020survey,wang2021robust,huang2022self,xu2019video,sabokrou2020deep}. The focus of this paper is on industrial defect detection~\cite{bergmann2019mvtec,jezek2021deep}, covering both image-level AD and pixel-level anomaly localization. The main challenge of unsupervised AD, compared to supervised approaches, lies in the absence of training annotations, be it at the image-level and pixel-level. Models are expected to be trained solely with normal instances, excluding anomalies. Unsupervised AD research can be broadly categorized into two tracks: one-class classification based AD approaches and self-supervised learning-based AD approaches.

\textbf{One-class classification-based AD approaches} are engineered to detect anomalies by developing a model of normal data instances and then assessing whether new instances align with this normative framework. These strategies presuppose that normal data can be encapsulated by compact models~\cite{Yamanishi2000On,Xu2012Robust,maziarka2021oneflow}, which anomalies do not fit. To more accurately define the distribution of normal data, conventional methods implement a range of statistical techniques~\cite{Eskin2000Anomaly,scholkopf2001estimating,Rahmani2017Coherence,ruff2018deep,kurt2020real}. OC-SVM~\cite{scholkopf2001estimating} utilizes a kernel function to project features into a space where normal instances are positively valued. Deep SVDD~\cite{ruff2018deep,yi2020patch} adopts a similar framework but integrates a deep convolutional neural network to reduce the volume of a hyper-sphere encompassing normal data representations. 

Recent advances in one-class classification for AD increasingly rely on feature embeddings from pre-trained networks, leveraging these as robust indicators of normalcy. Various statistical estimators of normal distributions~\cite{defard2021padim,orthoad,patchcore} help pinpoint significant deviations in feature behavior, flagging them as anomalies. Measures like the Mahalanobis distance~\cite{defard2021padim} are frequently used to calculate anomaly scores. Techniques such as those in \cite{defard2021padim,patchcore,cflow} have demonstrated notable success by utilizing models pretrained on the ImageNet dataset~\cite{imagenet}. However, solely relying on ImageNet pre-training has its limitations. This paper introduces feature registration as a novel proxy task that enhances feature aggregation and generalization capabilities, thus refining AD performance.
 
\textbf{Self-supervised learning-based AD approaches} leverage proxy tasks for model training through self-supervision~\cite{doersch2015unsupervised,noroozi2016unsupervised,gidaris2018unsupervised,golan2018deep,ARNet,bergmann2020uninformed,GP,luo2021future,wang20223}. Image reconstruction is a popular proxy, widely applied across data types such as images~\cite{xia2015learning,schlegl2017unsupervised,zong2018deep,Sabokrou2018Adversarially,ganomaly,gong2019memorizing,huang2022esad,metaformer,zhou2021memorizing,huang2022ssm,lo2022adversarially}, videos~\cite{luo2019video,georgescu2021background}, graphs~\cite{ahmed2021graph}, and time series~\cite{li2020anomaly}. These models, trained on normal data, identify anomalies through significant reconstruction errors. Alternative approaches include GeoTrans, which uses geometric image transformations to create a self-labeled dataset, employing transformation classification as the proxy task~\cite{golan2018deep,gidaris2018unsupervised}. CutPaste~\cite{cutpaste} introduces localized data augmentations for detecting small defects by learning to classify these modifications. ARNet~\cite{ARNet} enhances pixel-level semantic features through an image restoration framework, utilizing transformations to learn from corrupted inputs. Other methods like FYD~\cite{focus} employ non-contrastive learning as a proxy task, using coarse-to-fine alignment to extract discriminative features from normal images for a more compact normal sample distribution. Knowledge distillation is used in approaches like US~\cite{bergmann2020uninformed} and MKD~\cite{MKD}, where the student network learns by regressing to a teacher network's features, typically an ImageNet pre-trained model. Anomalies are detected when there is a noticeable difference between the student's and teacher's output features. Distinct from these, this paper introduces registration as a self-supervised task, specifically tailored for the FSAD scenario, which relies on only a few normal samples.

\subsection{Few-Shot Learning}
FSL addresses the task of adapting a model to novel categories with limited available images. FSL methods can be broadly categorized into metric learning~\cite{snell2017prototypical,guan2023query,willes2022bayesian}, generation~\cite{yang2021bridging,rewatbowornwong2022repurposing,yang2021free}, and optimization~\cite{li2023knowledge,finn2017model,jung2020few} approaches. Metric learning methods aim to structure the feature space to find the closest distance between test images and known classes. Generation methods involve designing generative models to produce novel class images, thereby bridging the few-shot gap. Optimization methods identify commonalities among different classes and use optimization models for novel classes based on these commonalities. In this paper, \texttt{CAReg} is based on metric learning to address the challenge of FSAD. The challenge is to detect image anomalies for novel categories that are unseen in the training dataset, using only a few normal samples. The key point is to enable the model to be category-agnostic, and this ability is provided by the proposed registration technique, allowing generalization across categories. Unlike previous work on FSL, training data and support sets contain only positive (normal) examples without any negative (anomalous) samples.

\subsection{Few-Shot Anomaly Detection} 
FSAD in industrial AD scenarios~\cite{lee2023few,gu2023prototype,li2024promptad} leverages a limited number of normal samples as a support set to detect and localize anomalies in unseen categories. This sophisticated task goes beyond simple AD to include fine-grained defect localization, presenting greater challenges than the broader, image-level decisions typical in traditional one-class classification~\cite{meta5} and out-of-distribution detection~\cite{sun2021react,sun2022out}.

Traditional FSAD approaches, such as TDG~\cite{TDG} and DiffNet~\cite{DiffNet}, often use a one-model-per-category strategy. TDG uses a hierarchical generative model to create multi-scale patch features, improving the diversity of the normal images in the support set through various transformations. In contrast, DiffNet extracts descriptive features at multiple scales using a pre-trained model and employs a normalizing flow to estimate anomaly distributions from a few support images.

Recent developments in FSAD are shifting towards category-agnostic techniques that enhance feature robustness and generalization without needing parameter fine-tuning. For example, RegAD~\cite{regad} uses a category-agnostic strategy to perform AD and localization in new categories. Building on this, RFR~\cite{lee2023few} uses adversarial loss, common in domain adaptation, to align feature distributions between source and target domains, improving the model's generalization capabilities. PACKD~\cite{gu2023prototype} further enhances generalization by integrating category-specific information from a teacher network into a student network, guiding the learning process with prototypes generated from few-shot normal samples. PromptAD~\cite{li2024promptad} introduces a dual-branch framework that uses prior knowledge about abnormal behaviors through text prompts, leveraging multimodal data to address visual information gaps and enhance AD performance. This paper sets itself apart by introducing a category-agnostic registration mechanism to train the model's backbone, transforming it into a flexible, plug-and-play tool that bolsters feature robustness and generalization for FSAD. This novel approach signifies a substantial advancement toward more adaptable and broadly applicable AD methods in industrial settings.
\section{Problem Formulation}\label{sec:formulation}

The training set, $\mathcal{T}_{train}=\{\mathcal{T}_{1}, \cdots, \mathcal{T}_{n}\}$, consists of only normal samples from $n$ categories, where $\mathcal{T}_{i}$ $(i=1, \cdots, n)$ is the sub-set for the $i$-th class. The test sample, consists of a test image $I^{\text{test}}$ from the $t$-th category ($t \in \{n+1, \cdots N$\}, $N>n$) and its corresponding support set $\mathcal{S}_{t}$ containing $K$ normal samples from the $t$-th category. Please note that at test-time, the model deals with novel categories unseen in the train data. For FSAD, we aim to use only a few normal images to detect anomalies in a test sample of novel categories, thus $K$ is set to a small number like 2, 4 or 8 in this paper.

The key challenges can be summarized as: (i) $\mathcal{T}_{train}$ only includes normal images without any image-level or pixel-level annotations, (ii) the test image categories and the training image categories are completely disjoint, and (iii) each test category only provides a few normal images.

\section{Category-Agnostic Registration Training}\label{sec:pretrain}

This section describes how to train a category-agnostic image encoder leveraging registration as a proxy task. For each training batch, we randomly sample a training image $I_a$ together with a support set of $k$ other images $B=\{I_b\}_{b=1}^k$ taken from the same category as $I_a$. Sec.~\ref{sec:stn} presents the architecture of the category-agnostic image encoder, with a spatial transformation module inserted into the convolutional residual blocks to enable registration. Sec.~\ref{sec:frm} presents the feature registration module.

\begin{figure*}[t]
\centering
\includegraphics[width=1.0\textwidth]{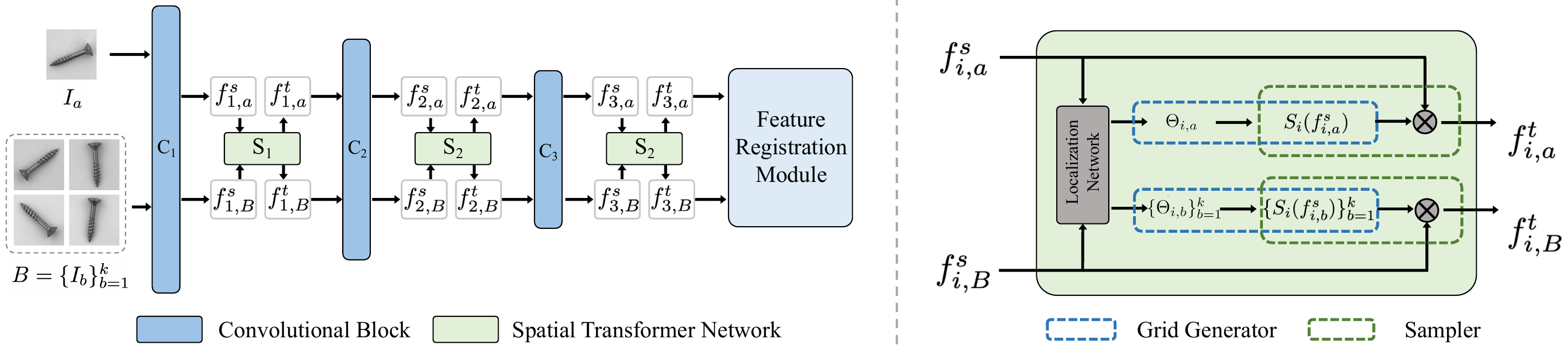}
\caption{(Left) An overview of the architecture of the proposed category-agnostic registration (\texttt{CAReg}) network. Given a train image and a set of support images, features are extracted by three convolutional residual blocks ($C_1$, $C_2$, and $C_3$), each followed by a spatial transformation module ($S_1$, $S_2$, and $S_3$). A feature registration module is leveraged and supervised by a registration loss (Fig.~\ref{img:RegAD_frm}). (Right) A spatial transformation module, containing a localization network and a differentiable sampler, is used to learn the mappings, enabling the model to transform features to facilitate feature registration.}
\label{img:RegAD}
\end{figure*}

\subsection{Architecture of the Category-Agnostic Encoder}\label{sec:stn}
As shown in Fig.~\ref{img:RegAD}, a residual convolutional network~\cite{he2016deep} is used as the feature extractor. Following~\cite{defard2021padim,focus}, only the first three convolutional residual blocks of ResNet, \emph{i.e.}, $C_1$, $C_2$, and $C_3$, are kept, while the last convolution block is discarded to ensure the output features retain sufficient spatial information. To enable the alignment of images with different orientation, a spatial transformation module employing a spatial transformer network (STN)~\cite{STN} is inserted into each convolutional residual block. Specifically, as shown in Fig.~\ref{img:RegAD}, given a pair of input features $f_{i,a}^s$ and $f_{i,B}^s=\{f_{i,b}^s\}_{b=1}^k$ extracted by the convolutional residual block $C_i$ for the training image $I_a$ and the set of support images $B$, a localization network is leveraged to generate affine transformation matrices $\Theta_{i,a}$ and $\{\Theta_{i,b}\}_{b=1}^k$ for feature registration. To perform a warping of the input features, the point-wise transformation is:
\begin{equation}
\begin{pmatrix}
x_i^s \\ y_i^s
\end{pmatrix}
= S_i(f_{i}^s)
=\Theta_i\begin{pmatrix}
x_{i}^t \\ y_{i}^t \\ 1
\end{pmatrix}
=\begin{bmatrix}
\theta_{11} & \theta_{12} & \theta_{13} \\
\theta_{21} & \theta_{22} & \theta_{23} \\
\end{bmatrix} 
\begin{pmatrix}
x_{i}^t \\ y_{i}^t \\ 1
\end{pmatrix},
\end{equation}
where $(x_i^t,y_i^t)$ are the target coordinates of the output feature $f_i^t$, $(x_i^s,y_i^s)$ are the source coordinates of the same point in the input feature $f_i^s$ and $\Theta_i$ is the affine transformation matrix set. A grid generator and a differentiable sampler~\cite{STN} are then used to generate the corresponding output features, $f_{i,a}^t$ and $f_{i,B}^t$. The architecture of the localization network is the same as that used in~\cite{STN}. Overall, for output features of the convolutional block $C_i$, the module $S_i$, containing a localization network, a grid generator, and a differentiable sampler, are used to learn the mappings. 

\begin{figure}[t]
\centering
\includegraphics[width=0.45\textwidth, height=6.3cm]{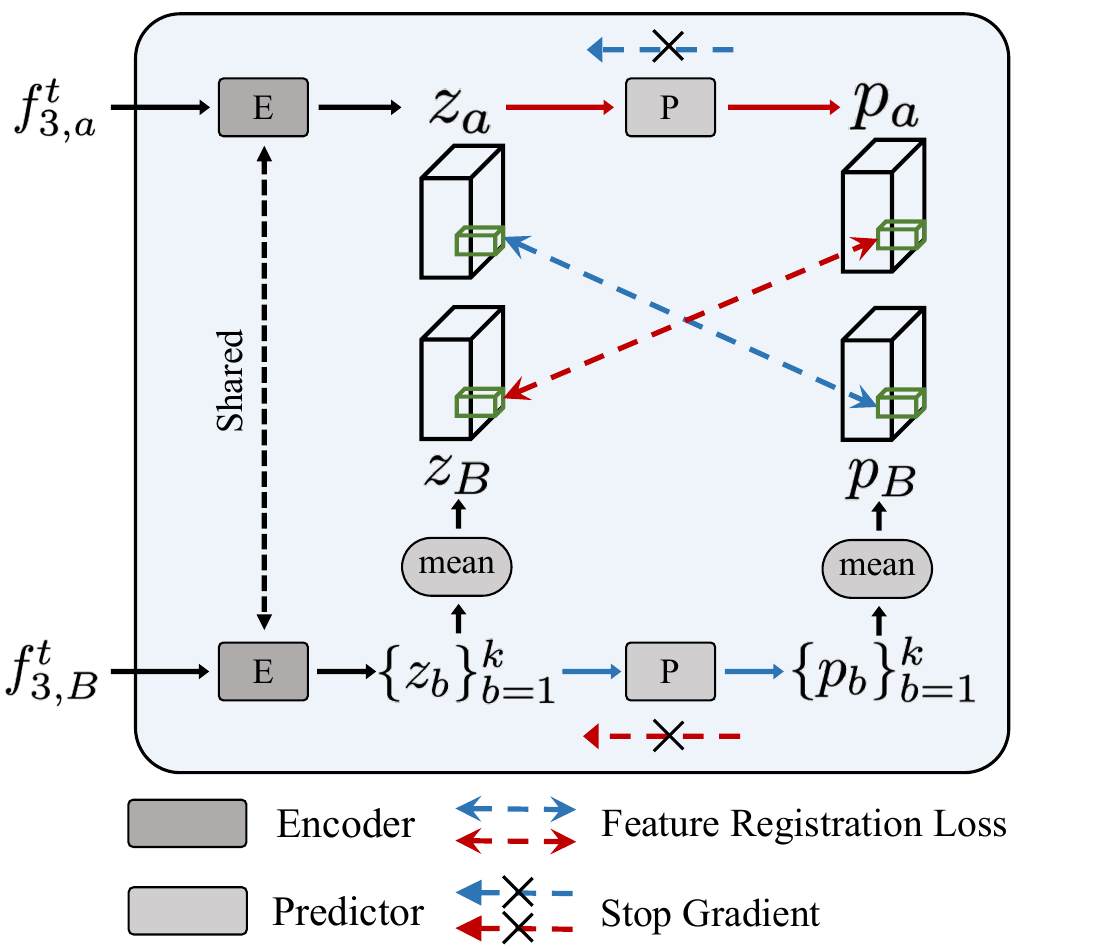}
\caption{The model architecture of the feature registration module. Given paired registered features, the parameter-shared encoder and predictor are leveraged and supervised by a registration loss.}
\label{img:RegAD_frm}
\end{figure}

\begin{figure*}[t]
\centering
\includegraphics[width=1.0\textwidth]{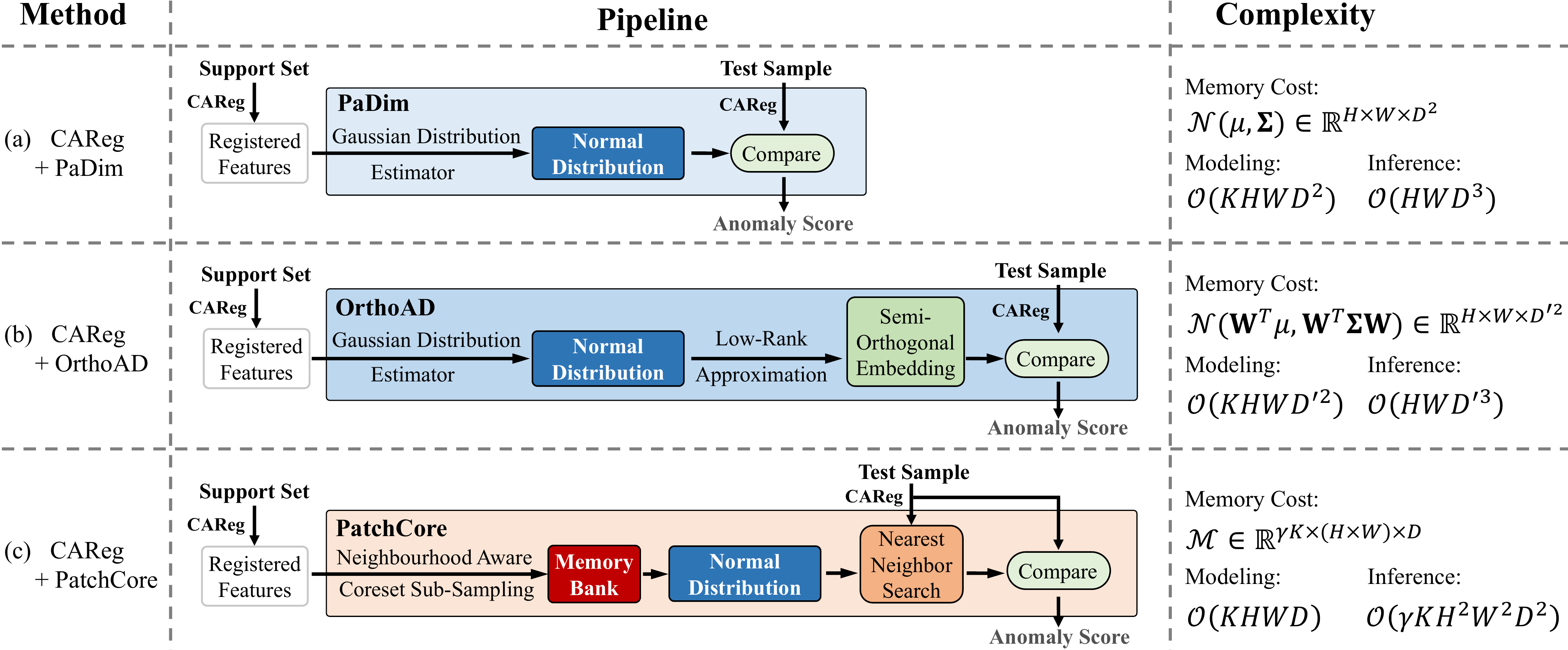}
\caption{AD methods and their corresponding memory cost, distribution modeling complexity, and inference complexity. Methods are shown by combining registration trained features extracted by \texttt{CAReg} and three statistical-based normal distribution estimators: (a) PaDim~\cite{defard2021padim}, (b) OrthoAD~\cite{orthoad}, and (c) PatchCore~\cite{patchcore}. Symbols used in the complexity equations include: $D$ is the sum of the channel dimensions for the three STN outputs, K is the size of the given support set, $D'$ is a constant and $D' \ll D$, and $\gamma$ denotes the proportion of the original memory bank that has been sampled. Other symbols are explained in the text.}
\label{img:anomalyheads}
\end{figure*}

\subsection{Feature Registration Module}\label{sec:frm}
Given paired extracted features $f_{3,a}^t$ from the training image and $f_{3,B}^t=\{f_{3,b}^t\}_{b=1}^k$ from the support image set as the final transformation outputs, we leverage a Siamese network~\cite{bromley1993signature} for feature registration, as shown in Fig.~\ref{img:RegAD_frm}. The architecture employs a parameter-sharing encoder, $E$, applied on multiple inputs, followed by a prediction head $P$. Training employs the method used by SimSiam~\cite{chen2021exploring} to avoid the collapsing problem when optimized without negative pairs. Denoting $p_a\triangleq P(E(f_{3,a}^t))$ and $z_B\triangleq \textit{mean}(E(\{f_{3,b}^t\}_{b=1}^k))$ where $\textit{mean}(\cdot)$ is an averaging operation on samples, a negative cosine similarity loss is applied:
\begin{equation}
    \mathcal{D}(p_a,z_B)=-\frac{p_a}{||p_a||_2}\cdot \frac{z_B}{||z_B||_2},
\end{equation}
where $||\cdot||_2$ is the $L_2$ norm. Similarly, for features $p_B\triangleq \textit{mean}(P(E(\{f_{3,b}^t\}_{b=1}^k)))$ and $z_a\triangleq E(f_{3,a}^t)$, a symmetrical similarity loss is applied:
\begin{equation}
    \mathcal{D}(p_B,z_a)=-\frac{p_B}{||p_B||_2}\cdot \frac{z_a}{||z_a||_2}.
\end{equation}
Different from~\cite{regad} where only one single image is used as the reference, to improve the robustness, $z_B$ and $p_B$ are accumulated features (AF) produced by accumulating the features from multiple references. To prevent the collapsing solutions, once the prediction head $P$ is applied on one branch, as shown in Fig.~\ref{img:RegAD_frm}, a stop-gradient operation is applied to the other branch. Finally, a symmetrized feature registration loss is defined as:
\begin{equation}\label{eq:loss}
    \mathcal{L}=\frac{1}{2}(\mathcal{D}(p_a,\texttt{sg}(z_B))+\mathcal{D}(p_B,\texttt{sg}(z_a))),
\end{equation}
where $\texttt{sg}(\cdot)$ is the stop gradient operation. 
Capitalizing on the inherent robustness and superior performance of feature-level comparisons compared to direct image-level comparisons~\cite{defard2021padim}, the registration in this study is carried out at the feature level to directly acquire the registered features, instead of pixel-by-pixel image-level registration approaches. 

\noindent\textbf{Discussion.}
For the model architecture of the feature registration network, besides adopting the first three convolutional blocks of ResNet and removing the global average pooling, in the feature registration module, a convolutional encoder and predictor architecture is used to replace the MLP architecture in SimSiam~\cite{chen2021exploring}. As a result, the features retain relatively complete spatial information. For the proposed feature registration loss function, Eq.~\eqref{eq:loss}, we average the cosine similarity scores at every spatial pixel. Since anomaly score maps at the pixel level are required for anomaly localization, it is very important to retain the spatial information in the final features. Different from~\cite{chen2021exploring,focus} where the same images from a mini-batch or two augmentations of one image are used as inputs for contrastive representation learning, the proposed method leverages different images as inputs and learns transformation matrices through STN for feature registration. 

\section{Anomaly Detection}\label{sec:ad}
To detect anomalies, we begin by estimating the normal distribution of the target category using a support set consisting of a few normal images (Sec.~\ref{sec:distribution}). As the task of estimating a distribution from a sample set has been extensively studied in statistics, we adopt recent statistical-based estimators for this purpose. Considering the challenge posed by the limited size of the support set, we propose an augmentation auto-selection module (Sec.~\ref{sec:aug}) to choose an optimal set of data augmentations that can be applied to the support images during the estimation process. Then, we compare the test sample with the estimated normal distribution to produce an anomaly score (Sec.~\ref{sec:infer}). This anomaly score indicates the degree of deviation of the test sample from the estimated normal distribution and serves as a measure for AD.

\subsection{Normal Distribution Estimation}\label{sec:distribution}
With the assumption that the registration can generalize across different categories without any parameter fine-tuning, the registration network is applied directly to the support set $\mathcal{S}$ for the target category. Suppose an image is divided into a grid of $(i,j)\in [1,W]\times [1,H]$ positions where $W\times H$ is the resolution of the registered features. The outputs of the three STN modules are concatenated at each corresponding patch position $(i,j)$, with upsampling used to match their sizes, to produce aggregated features $f_{ij}$. $\mathcal{F}_{reg} = \{ f_{ij}^k,k\in [1,K],i \in [1,W], j\in [1,H]\}$ denotes the set of registered features for the $K$ support images. The normal distribution $\mathcal{D}_{norm}$ of the target category is estimated with a statistical-based estimator 
\begin{equation}
\mathcal{D}_{norm} \triangleq \mathcal{E}_{norm}(\mathcal{F}_{reg}). 
\end{equation}

Below we briefly present three popular estimators that may be adopted. 
These estimators are illustrated in Fig.~\ref{img:anomalyheads}.

\begin{itemize}[leftmargin=*]
\item 
\textbf{PaDim}: The PaDim estimator~\cite{defard2021padim} adopts the multivariate Gaussian distribution $\mathcal{N}(\mu_{ij}, \boldsymbol{\Sigma}_{ij})$ as the normal distribution $\mathcal{D}_{norm,ij}$, where $\mu_{ij}$ and $\boldsymbol{\Sigma}_{ij}$ are the mean and covariance of features $\mathcal{F}_{reg}$ corresponding to the patch position $i,j$. To make the sample covariance matrix full rank and invertible, a regularization term $\epsilon I$ is introduced to the covariance: $\boldsymbol{\Sigma}_{ij}=\frac{1}{K-1} \sum_{k=1}^{K}(f_{ij}^k-\mu_{i j})(f_{ij}^k-\mu_{i j})^{\mathrm{T}}+\epsilon I$. Each possible patch position is associated with a multivariate Gaussian distribution $\mathcal{D}_{norm}=(\mathcal{D}_{norm,ij})_{1\leqslant i\leqslant W, 1\leqslant j\leqslant H}$.

\item
\textbf{OrthoAD}: To reduce the memory cost and accelerate the calculation of the PaDim estimator, OrthoAD~\cite{orthoad} leverages a low-rank embedding of input features. With a semi-orthogonal matrix $\mathbf{W} \in \mathbb{R}^{D \times D'}$~\cite{mezzadri2006generate}, where $D'$ is a constant and $D' \ll D$, the estimated normal distribution could have a low-rank approximation, where the low-rank sample mean is replaced with $\mathbf{W}^T \mu_{ij} \in \mathbb{R}^{D'}$, and the low-rank sample covariance is replaced with $\mathbf{W}^T \boldsymbol{\Sigma}_{ij} \mathbf{W} \in \mathbb{R}^{D' \times D'}$.

\item
\textbf{PatchCore}: To further reduce the memory costs while maintaining nominal information at
test time, PatchCore~\cite{patchcore} uses the coreset sampling~\cite{sener2018active} to build a maximally representative memory bank $\mathcal{M}$, which stores the neighbourhood-aware features from all normal samples. The normal distribution $\mathcal{D}_{norm}$ is approximated with $\texttt{Coreset}\left(\mathcal{F}_{reg}\right)$ where $\texttt{Coreset}(\cdot)$ is an iterative greedy approximation~\cite{sener2018active} to reduce the number of memory items (detailed in Algorithm 1 in the appendix).
\end{itemize}

\noindent \textbf{Complexity Analysis}: The complexity analysis of the above three distribution estimator is summarized in Fig.~\ref{img:anomalyheads}. It can be seen that compared with PaDim, OrthoAD significantly reduces the memory cost due to its low-rank approximation for the normal distribution $\mathcal{D}_{norm}$. The complexity of the distribution modeling of PatchCore is the minimum modeling complexity among the three estimators.

\subsection{Augmentation Selection Module}\label{sec:aug}
AD approaches employ data augmentation as an important tool to expand the dataset, especially in the FSAD scenario~\cite{TDG,DiffNet}. However, the impact of augmentation, specifically, where and what data augmentation to use, has not been fully explored. Instead of simply applying the data augmentations on both the support and test images, this paper shows that augmentations, by expanding the support set, play a very important role in the normal distribution estimation process. We only augment the support set, which also reduces the computational cost. Possible combinations of the augmentations produce a larger augmented support set. We conduct the normal distribution estimation on such an augmented support set. 

In addition to using augmentations consistent with those used by RegAD~\cite{regad}, we further propose a Wasserstein distance-based data augmentation selection mechanism to flexibly choose the most suitable augmentations for each category. According to the different visual characteristics among categories, it is meaningful to remove the augmentations that may damage the main properties of images. Given the support set $\mathcal{S}$ and its augmented set $\mathcal{S}_c$ with the $c$-th augmentation, denote the normal distributions estimated by the estimator be $\mathcal{N}(\mu_{ij}, \boldsymbol{\Sigma}_{ij})$ and $\mathcal{N}(\mu_{c,ij}, \boldsymbol{\Sigma}_{c,ij})$ for $\mathcal{S}$ and $\mathcal{S}_c$ at position $(i,j)$, respectively. The Wasserstein distance~\cite{givens1984class} at position $(i,j)$ between two distributions is defined as:
\begin{equation}\label{eq:wd}
    \mathcal{W}_{c,ij}
    = ||\mu-\mu_c||^2_2 + Tr(\boldsymbol{\Sigma}+\boldsymbol{\Sigma}_c-2(\boldsymbol{\Sigma}^{\frac{1}{2}}\boldsymbol{\Sigma}_c\boldsymbol{\Sigma}^{\frac{1}{2}})^{\frac{1}{2}}),
\end{equation}
where we omit the subscript $(\cdot)_{ij}$ for all variables in the right-hand side of Eq.~\eqref{eq:wd} to simplify the notation. To focus on the foreground, the Wasserstein distances in Eq.~\eqref{eq:wd} are re-weighted and then summed, overall different feature patches at positions $(i,j)$:
\begin{equation}
    \mathcal{W}_{c}' = \sum_{ij}(||\mu_{ij}-\mu_{c,ij}||_2 \cdot \mathcal{W}_{c,ij}), 
\end{equation}
where $\mathcal{W}_{c,ij}$ is the Wasserstein distance score at patch position $(i,j)$ obtained from Eq.~\eqref{eq:wd}, and $\mathcal{W}_{c}'$ represents the weighted sum of the Wasserstein distance scores over all patch positions. With an assumption that the $L_2$ difference of the sample mean on the foreground is larger, through the re-weighting, the Wasserstein distance of the background area is expected to be small, while the foreground related to the objects is highlighted. Given a set of Wasserstein distances $\{\mathcal{W}_{c}'\}_{c=1}^{n}$ for $n$ kinds of augmentations, denote a threshold as the average of the Wasserstein distances, $\delta \triangleq \frac{1}{n}\sum_c \mathcal{W}_{c}'$, and then augmentations with higher Wasserstein distance than this threshold are removed from the augmentation set. The remaining augmentations are combined to jointly augment the support set. Note that the proposed augmentation selection pipeline is only used in the normal distribution estimation stage. As no data augmentation is used in the inference stage, the efficiency of the inference is not affected.

\subsection{Anomaly Scoring at Inference}\label{sec:infer}
During inference, we compare the registered features of the test image to its corresponding normal distribution to detect anomalies: test samples out of the normal distribution are considered anomalies. Given a test image $I^{\text{test}}$ and the estimated normal distribution  $\mathcal{D}_{norm}$, denote $f_{ij}$ as the registered feature of $I^{\text{test}}$ at the patch position $(i,j)$, the anomaly score of the patch at position $(i,j)$ is formulated as:
\begin{equation}\label{eq:distance}
    \mathit{d}_{ij}=Dist(f_{ij}, \mathcal{D}_{norm}),
\end{equation}
where $Dist(\cdot, \cdot)$ is a distance function between the feature and its corresponding normal distribution.
The matrix of distances $\mathit{d}=(\mathit{d}_{ij})_{1\leqslant i\leqslant W, 1\leqslant j\leqslant H}$ forms an anomaly map, with anomalous areas indicated with high scores. For anomaly localization, three inverse transform matrices, corresponding to the three STN modules, are applied to re-match the regions of spatial-transformed features and the original images, thus getting the final anomaly score map $\mathit{d}_{final}$. The image-level anomaly score is the maximum of this anomaly 
map.

To ensure consistency between the anomaly scoring functions and the distribution estimators, a corresponding distance function is defined for each estimator. 
\begin{itemize}
[leftmargin=*]
    \item \textbf{PaDim}: A Mahalanobis distance is adopted:
\begin{equation}\label{eq:padim}
    Dist\left(f_{ij}, \mathcal{D}_{norm}\right)=\sqrt{\left(f_{ij}-\mu_{ij}\right)^{T} \boldsymbol{\Sigma}_{ij}^{-1}\left(f_{ij}-\mu_{ij}\right)},
\end{equation}
where $\mu_{ij}$ and $\boldsymbol{\Sigma}_{ij}$ are the mean and covariance of the normal distribution $\mathcal{D}_{norm}$ at position $(i,j)$. Note that in Eq.~\eqref{eq:padim}, calculating the inverses of covariance matrices $\boldsymbol{\Sigma}_{ij}^{-1}\in \mathbb{R}^{D \times D}$ has a computational complexity of $\mathcal{O}\left(H W D^3\right)$, which prohibits efficient computation with large dimensional multi-scale features.
\item \textbf{OrthoAD}: $Dist(\cdot, \cdot)$ is defined the same as in Eq.~\eqref{eq:padim} but in low-rank, where the feature $f_{ij}$, the sample mean $\mu_{ij}$, and the sample covariance $\boldsymbol{\Sigma}_{ij}$ are replaced with their low-rank correspondence, \emph{i.e.}, $\mathbf{W}^Tf_{i j}$, $\mathbf{W}^T\mu_{i j}$, and $\left(\mathbf{W}^T\boldsymbol{\Sigma}_{i j}\mathbf{W}\right)^{-1}$, respectively. The computational complexity is thus cubically reduced to $\mathcal{O}\left(H W D'^3\right)$, where $D' \ll D$.
\item \textbf{PatchCore}: 
With the normal distribution formulated as a memory bank under the PatchCore estimator, the distance function $Dist(\cdot, \cdot)$ is defined as the minimum $L_2$ distance between the patch feature of the test image and its respective nearest neighbour memory $m^*$ in the normal patches feature bank $\mathcal{D}_{norm}$,
\begin{equation}\label{eq:patchcore}
Dist\left(f_{ij}, \mathcal{D}_{norm}\right)=\left\|f_{ij}-m^*\right\|_2.
\end{equation}
where a nearest neighbor search is used to get the optimal $m^*=\arg \min _{m \in \mathcal{D}_{norm}}\left\|f_{ij}-m\right\|_2$. A re-weight function~\cite{patchcore} is applied to Eq.~\eqref{eq:patchcore} to obtain the final anomaly map considering neighbor patches. The computational complexity is $\mathcal{O}\left(\gamma K H^2 W^2 D^2\right)$ due to the need to traverse $\gamma K H W$ memory items in $\mathcal{D}_{norm}$.
\end{itemize}
\vspace{-3mm}
\section{Experiments}\label{sec:experiment}

\subsection{Datasets and Experimental Setting}

\noindent \textbf{MVTec AD dataset~\cite{bergmann2019mvtec}.} MVTec is a challenging industrial defect detection dataset, comprising 15 categories with a total of 3629 training and 1725 testing images. Following the traditional AD setting, only normal images, without any defects, are provided during training. The test set contains images showing various kinds of defects and defect-free (normal) images. 73 different anomalous types are given, on average five per category. Five categories in this dataset cover different types of random (\emph{i.e.}, leather, tile, and wood) or regular (\emph{i.e.}, carpet and grid) textures, while the other ten categories are for various types of objects. Pixel-level ground truth labels are provided for the defective image regions. 

\noindent \textbf{MPDD AD dataset~\cite{jezek2021deep}.} MPDD provides images of six categories captured during painted metal part fabrication. In contrast to MVTec, it focuses specifically on defect detection under varying viewing conditions, such as different spatial orientations and positions, different lighting conditions, and non-homogeneous backgrounds. Thus, it is more challenging and more complex than the MVTec. Similar to MVTec, various types of defects are present in the test images, which cover a wide range of scenarios that can occur in the painting industry and metal fabrication. We summarize the differences between the two industrial AD datasets in Table~\ref{tal:datasetsum}.

\begin{table}[tp]
\centering
\small
\caption{Summary of the key differences between the two industrial AD datasets used for benchmarking.}\label{tal:datasetsum}
\scalebox{0.92}{
\setlength{\tabcolsep}{2.5pt}{
\begin{tabular}{L{2.5cm}|L{2.6cm}|L{3.2cm}}
\toprule
dataset & MVTec & MPDD\\
\hline
categories & \makecell[l]{10 objects and\\ 5 texture surfaces} & 6 classes of metal parts\\
\hline
position & centered & \makecell[l]{positioned and\\ rotated differently}\\
\hline
background & \makecell[l]{homogeneous\\monochromatic} & reflections or shadows\\
\hline
number of objects & one & many\\
\hline
motion blur & no & yes\\
\hline
variability & low & high\\
\bottomrule
\end{tabular}}
}
\end{table}

\begin{table*}[t]
\centering
\caption{Comparison of \texttt{CAReg} with two state-of-the-art FSAD methods on MVTec and MPDD with $K=\{2,4,8\}$. Results of the three normal distribution estimation methods, PaDiM, OrthoAD (OAD), and PatchCore (PC), are also reported as baselines.
The AUC (in \%) averaged over 10 runs is reported, with the best result in each row marked in bold.}
\label{tal:sum}
\vspace{-4pt}
\small
\scalebox{0.97}{
\setlength{\tabcolsep}{1.0pt}{
\begin{tabular}{C{1.2cm}C{1.0cm}|C{1.1cm}>{\columncolor{gray!15}}C{1.65cm}C{1.2cm}>{\columncolor{gray!15}}C{1.65cm}||C{1.3cm}>{\columncolor{gray!15}}C{1.8cm}C{1.4cm}>{\columncolor{gray!15}}C{1.8cm}C{1.4cm}>{\columncolor{gray!15}}C{1.8cm}}
\toprule
Dataset & Shots &
\makecell[c]{TDG\\\cite{TDG}} & \makecell[c]{TDG+\\\cite{TDG}} & \makecell[c]{DiffNet\\\cite{DiffNet}} & \makecell[c]{DiffNet+\\\cite{DiffNet}} & \makecell[c]{PaDim\\\cite{defard2021padim}} & \makecell[c]{PaDim\\+CAReg}  & \makecell[c]{OAD\\\cite{orthoad}} & \makecell[c]{OAD\\+CAReg} & \makecell[c]{PC\\\cite{patchcore}} & \makecell[c]{PC\\+CAReg}\\
\cmidrule(lr){1-12}  
\multirow{3}{*}{MVTec} & K=2     & 71.2 & 73.2 (+2.0) & 80.5 & 80.6 (+0.1) & 74.7 & 85.5 (+10.8) & 76.7 & 83.6 (+6.9) & 85.8 & \textbf{90.2} (+4.4)\\
& K=4     & 72.7 & 74.4 (+1.7) & 80.8 & 81.3 (+0.5) & 78.0 & 89.2 (+11.2) & 77.6 & 86.9 (+9.3) & 88.8 & \textbf{93.7} (+4.9)\\
& K=8     & 75.2 & 76.6 (+1.4) & 82.9 & 83.2 (+0.3) & 80.5 & 91.2 (+10.7) & 79.8 & 89.2 (+9.4) & 91.3 & \textbf{95.1} (+3.8)\\
\cmidrule(lr){1-12}   
\multirow{3}{*}{MPDD} & K=2     & 57.3 & 60.3 (+3.0) & 58.4 & 60.2 (+1.8) & 49.6 & 64.8 (+15.2) & 48.6 & 57.8 (+9.2) & 58.5 & \textbf{67.3} (+8.8)\\
& K=4     & 60.4 & 63.5 (+3.1) & 61.2 & 63.3 (+2.1) & 53.7 & 67.8 (+14.1) & 50.3 & 60.3 (+10.0) & 64.5 & \textbf{72.4} (+7.9)\\
& K=8     & 64.4 & 68.2 (+3.8) & 66.5 & 68.5 (+2.0) & 55.5 & 72.5 (+17.0) & 53.0 & 70.1 (+17.1) & 69.9 & \textbf{77.3} (+7.4)\\
\bottomrule
\end{tabular}
}}
\end{table*}

\begin{table*}[t]
\centering
\caption{Comparison of \texttt{CAReg} with its corresponding baseline 
normal distribution estimation methods on MVTec with $K=\{2,4,8\}$. AUC in \% averaged over 10 runs for each category and the standard deviation (SD) of PC and PC + \texttt{CAReg} in percentage (\% is omitted) over these 10 runs (10 different support sets) are provided, together with a macro-average score over all categories. For each pair, the best-performing method is marked in bold.}
\label{tal:mvtec}
\vspace{-4pt}
\scriptsize
\scalebox{0.96}{
\setlength{\tabcolsep}{0.65pt}{
\begin{tabular}{C{1.25cm}C{0.95cm}>{\columncolor{gray!15}}C{1.06cm}C{0.85cm}>{\columncolor{gray!15}}C{0.88cm}C{0.85cm}>{\columncolor{gray!15}}C{0.88cm}C{0.95cm}>{\columncolor{gray!15}}C{1.06cm}C{0.85cm}>{\columncolor{gray!15}}C{0.88cm}C{0.85cm}>{\columncolor{gray!15}}C{0.88cm}C{0.95cm}>{\columncolor{gray!15}}C{1.06cm}C{0.85cm}>{\columncolor{gray!15}}C{0.88cm}C{0.85cm}>{\columncolor{gray!15}}C{0.88cm}}
\toprule
\multirow{3}{*}{Category} & \multicolumn{6}{c}{K=2} & \multicolumn{6}{c}{K=4} & \multicolumn{6}{c}{K=8}\\
\cmidrule(lr){2-7} \cmidrule(lr){8-13} \cmidrule(lr){14-19}
& \makecell[c]{PaDim\\\cite{defard2021padim}} & \makecell[c]{PaDim+\\CAReg} & \makecell[c]{OAD\\\cite{orthoad}} & \makecell[c]{OAD+\\CAReg} & \makecell[c]{PC\\\cite{patchcore}} & \makecell[c]{PC+\\CAReg} & \makecell[c]{PaDim\\\cite{defard2021padim}} & \makecell[c]{PaDim+\\CAReg} & \makecell[c]{OAD\\\cite{orthoad}} & \makecell[c]{OAD+\\CAReg} & \makecell[c]{PC\\\cite{patchcore}} & \makecell[c]{PC+\\CAReg} & \makecell[c]{PaDim\\\cite{defard2021padim}} & \makecell[c]{PaDim+\\CAReg} & \makecell[c]{OAD\\\cite{orthoad}} & \makecell[c]{OAD+\\CAReg} & \makecell[c]{PC\\\cite{patchcore}} & \makecell[c]{PC+\\CAReg}\\
\cmidrule(lr){1-19}
Bottle     & 97.5 & \textbf{99.7} & \textbf{98.5} & 98.0 & \textbf{99.8} \tiny \textbf{0.5} & 99.6 \tiny \textbf{0.5} & 98.3 & \textbf{99.3} & 98.3 & \textbf{99.4} & 99.4 \tiny 0.2 & \textbf{99.9} \tiny \textbf{0.1} & 98.8 & \textbf{99.8} & 99.3 & \textbf{99.6} & \textbf{100} \tiny 0.2 & 99.9 \tiny \textbf{0.0} \\
Cable      & 52.1 & \textbf{69.8} & 57.2 & \textbf{63.1} & 86.2 \tiny 5.0 & \textbf{91.5} \tiny \textbf{2.8} & 58.9 & \textbf{82.9} & 55.5 & \textbf{67.4} & 88.9 \tiny 2.5 & \textbf{93.8} \tiny \textbf{2.0} & 61.5 & \textbf{81.5} & 60.6 & \textbf{74.1} & 88.8 \tiny 2.4 & \textbf{95.2} \tiny \textbf{1.2} \\
Capsule    & 56.9 & \textbf{68.6} & 67.3 & \textbf{80.1} & 66.2 \tiny \textbf{4.9} & \textbf{70.5} \tiny 5.7 & 65.4 & \textbf{77.3} & 69.4 & \textbf{85.2} & 76.1 \tiny \textbf{1.9} & \textbf{81.4} \tiny 3.2 & 71.7 & \textbf{78.4} & 71.0 & \textbf{86.8} & \textbf{88.7} \tiny \textbf{4.5} & 82.0 \tiny 5.5 \\
Carpet     & 90.9 & \textbf{96.7} & 95.1 & \textbf{98.5} & 99.0 \tiny \textbf{0.6} & \textbf{99.4} \tiny \textbf{0.6} & 92.8 & \textbf{97.9} & 96.6 & \textbf{98.5} & \textbf{99.1} \tiny 0.7 & 99.0 \tiny \textbf{0.4} & 93.9 & \textbf{98.6} & 97.3 & \textbf{99.3} & 99.1 \tiny 0.4 & \textbf{99.3} \tiny \textbf{0.2} \\
Grid       & 67.2 & \textbf{79.1} & 71.8 & \textbf{77.0} & 58.3 \tiny 5.7 & \textbf{79.9} \tiny \textbf{5.5} & 66.1 & \textbf{87.0} & 60.9 & \textbf{82.0} & 62.2 \tiny 8.4 & \textbf{89.8} \tiny \textbf{4.6} & 70.0 & \textbf{91.5} & 68.5 & \textbf{91.6} & 70.1 \tiny 4.9 & \textbf{91.0} \tiny \textbf{4.0} \\
Hazelnut   & \textbf{97.2} & 96.3 & 97.7 & \textbf{98.6} & 89.5 \tiny 5.5 & \textbf{97.9} \tiny \textbf{1.0} & \textbf{96.9} & 95.9 & 97.3 & \textbf{97.8} & 98.8 \tiny 4.4 & \textbf{99.4} \tiny \textbf{1.5} & 96.2 & \textbf{97.3} & \textbf{96.7} & 94.4 & 99.5 \tiny 0.9 & \textbf{99.8} \tiny \textbf{0.7} \\
Leather    & 86.6 & \textbf{100} & 99.7 & \textbf{99.8} & \textbf{100} \tiny 0.6 & \textbf{100} \tiny \textbf{0.2} & \textbf{100} & \textbf{100} & \textbf{99.9} & \textbf{99.9} & \textbf{100} \tiny 0.7 & \textbf{100} \tiny \textbf{0.2} & 92.7 & \textbf{100} & \textbf{99.9} & \textbf{99.9}  & \textbf{100} \tiny 0.5 & \textbf{100} \tiny \textbf{0.1} \\
Metal Nut  & 53.5 & \textbf{94.2} & 49.2 & \textbf{87.6} & 92.0 \tiny 4.5 & \textbf{97.7} \tiny \textbf{1.7} & 87.9 & \textbf{94.3} & 54.9 & \textbf{92.5} & 92.4 \tiny 3.1 & \textbf{98.2} \tiny \textbf{0.9} & 66.8 & \textbf{98.6} & 77.1 & \textbf{95.0} & 95.1 \tiny 2.4 & \textbf{98.1} \tiny \textbf{0.9} \\
Pill       & 48.3 & \textbf{66.1} & 58.5 & \textbf{72.4} & 83.7 \tiny 4.3 & \textbf{86.1} \tiny \textbf{3.1} & 61.1 & \textbf{74.0} & 58.4 & \textbf{78.8} & 84.3 \tiny 4.3 & \textbf{90.8} \tiny \textbf{3.0} & 53.4 & \textbf{77.8} & 47.2 & \textbf{80.9} & 89.3 \tiny 3.9 & \textbf{90.4} \tiny \textbf{1.3} \\
Screw      & 53.3 & \textbf{53.9} & \textbf{55.1} & 53.1 & 45.2 \tiny \textbf{5.5} & \textbf{62.2} \tiny 6.7 & 51.4 & \textbf{59.3} & 59.1 & \textbf{66.8} & 51.1 \tiny \textbf{4.1} & \textbf{73.7} \tiny 6.8 & 52.7 & \textbf{65.8} & 55.8 & \textbf{76.3} & 47.6 \tiny 5.3 & \textbf{80.9} \tiny \textbf{4.2} \\
Tile       & 81.1 & \textbf{98.9} & 74.5 & \textbf{97.0} & 98.9 \tiny 0.4 & \textbf{100} \tiny \textbf{0.1} & 55.5 & \textbf{98.2} & 86.2 & \textbf{97.0} & 99.3 \tiny 0.3 & \textbf{100} \tiny \textbf{0.1} & 86.5 & \textbf{99.6} & 88.8 & \textbf{96.9} & 99.2 \tiny 0.1  & \textbf{99.9} \tiny \textbf{0.0} \\
Toothbrush & \textbf{88.8} & 86.8 & \textbf{80.6} & \textbf{80.6} & \textbf{86.4} \tiny \textbf{4.7} & 83.0 \tiny 5.0 & \textbf{93.9} & 91.1 & 79.2 & \textbf{81.8} & \textbf{87.2} \tiny \textbf{2.9} & 86.6 \tiny 4.4 & \textbf{98.8} & 96.6 & \textbf{79.7} & \textbf{79.7} & \textbf{95.8} \tiny \textbf{1.4} & 94.8 \tiny 1.6 \\
Transistor & 74.6 & \textbf{82.2} & \textbf{75.7} & 70.7 & 85.0 \tiny 14.2 & \textbf{91.2} \tiny \textbf{6.5} & 78.3 & \textbf{85.5} & \textbf{77.9} & 75.8 & 97.1 \tiny 7.9 & \textbf{97.8} \tiny \textbf{4.8} & 85.2 & \textbf{90.3} & \textbf{78.1} & 76.4 & \textbf{100} \tiny \tiny \textbf{1.3} & 99.2 \tiny \textbf{1.3} \\
Wood       & 97.8 & \textbf{99.8} & 96.9 & \textbf{98.5} & 98.9 \tiny 1.0 & \textbf{99.0} \tiny \textbf{0.5} & 98.4 & \textbf{98.9} & 97.2 & \textbf{99.0} & 98.0 \tiny 1.0 & \textbf{98.6} \tiny \textbf{0.7} & 98.9 & \textbf{99.5} & 98.3 & \textbf{98.9} & 97.8 \tiny \textbf{0.4} & \textbf{98.6} \tiny 0.5 \\
Zipper     & 74.6 & \textbf{90.9} & 73.2 & \textbf{79.2} & \textbf{98.3} \tiny \textbf{0.9} & 95.2 \tiny 1.9 & 79.4 & \textbf{95.8} & 73.9 & \textbf{82.1} & \textbf{98.3} \tiny 1.4 & 96.5 \tiny \textbf{1.0} & 80.6 & \textbf{93.4} & 79.1 & \textbf{88.3} & \textbf{98.0} \tiny \tiny \textbf{1.1} & 96.8 \tiny \textbf{1.1} \\
\cmidrule(lr){1-19}
Average    & 74.7 & \textbf{85.5} & 76.7 & \textbf{83.6} & 85.8 \tiny 3.9 & \textbf{90.2} \tiny \textbf{2.8} & 78.0 & \textbf{89.2} & 77.6 & \textbf{86.9} & 88.8 \tiny 2.9 & \textbf{93.7} \tiny \textbf{2.2} & 80.5 & \textbf{91.2} & 79.8 & \textbf{89.2} & 91.3 \tiny 2.0 & \textbf{95.1} \tiny \textbf{1.5} \\
\bottomrule
\end{tabular}
}}
\end{table*}

\noindent\textbf{Experimental Setting.} The experiments for FSAD were conducted under the \textbf{leave-one-out setting}. In this configuration, a designated target category was chosen for testing, while the remaining categories in the dataset were employed for training. The aim of this approach was to gauge the AD performance when confronted with an \textbf{unseen category}, thereby assessing the potential of the model's capacity for generalization across all categories beyond the categories seen during the training. To create a demanding few-shot learning context, all data corresponding to the target category were excluded from the training set. During testing, only an exceedingly limited number of normal samples from the target category were made available in the support set. We conducted 10 separate experiments, each utilizing a distinct support set selected at random. After the selection, the support set for each experiment was fixed to maintain consistency and facilitate comparisons with state-of-the-art methods, yielding a variety of 10 outcomes. The reported performance metrics were the average results across these 10 trials.

\subsection{Competing Methods and Baselines} 
In this study, two state-of-the-art FSAD methods are considered as the baseline models, including TDG~\cite{TDG} and DiffNet~\cite{DiffNet}. Official source codes for these methods are used to train the category-dependent models. Additionally, the methods are extended to leverage data from multiple categories through a pre-training procedure, resulting in the methods TDG+ and DiffNet+. TDG+ uses data from multiple categories to pre-train the transformation classifier, using the vanilla AD method GeoTrans~\cite{golan2018deep}. DiffNet+ pre-trains the normalizing flow, which is the normal distribution estimator. RFR~\cite{lee2023few}, PACKD~\cite{gu2023prototype}, and PromptAD~\cite{li2024promptad} are also considered.

To demonstrate the effectiveness of \texttt{CAReg} over normal distribution estimators for FSAD, three state-of-the-art statistical-based normal distribution estimation methods, PaDiM~\cite{defard2021padim}, OrthoAD (OAD)~\cite{orthoad}, and PatchCore (PC)~\cite{patchcore}, are considered as the baselines.

\texttt{CAReg} is further compared with some state-of-the-art vanilla AD methods, including GANomaly~\cite{ganomaly}, ARNet~\cite{ARNet}, MKD~\cite{MKD}, CutPaste~\cite{cutpaste}, FYD~\cite{focus}, PaDiM~\cite{defard2021padim}, PatchCore~\cite{patchcore}, and CflowAD~\cite{cflow}. These methods serve as an upper bound on FSAD performance as they use the entire normal dataset for training, unlike FSAD methods that work with a few normal samples.

\vspace{-2mm}
\subsection{Evaluation Protocols and Model Configurations}
\noindent\textbf{Evaluation Protocols.} The area under the Receiver Operating Characteristic curve metric (AUC) is used to quantify the model performance. For AD and anomaly localization, AUC at the image-level and the pixel-level are considered, respectively. This metric is widely used for AD performance evaluation.

\noindent\textbf{Model Configuration and Training Details.} For the registration backbone, a ResNet-18~\cite{he2016deep} architecture pre-trained on the ImageNet is adopted. The feature registration module consists of an encoder and a predictor, similar to those used in previous work~\cite{chen2021exploring,focus}. The encoder includes three $1\times 1$ convolutional layers, while the predictor includes two $1\times 1$ convolutional layers. Pooling operations are removed from the ResNet backbone to retain spatial information in registered features. Models are trained on a single NVIDIA GTX 3090 GPU. Images are resized to have a resolution of $224 \times 224$ pixels. Standard image augmentations used in previous work~\cite{TDG,regad} are applied. The model parameters are updated for 50 epochs, with an initial learning rate of 0.0001, updated using a single cycle of the cosine learning rate scheduler. The batch size is set to 32, and the optimizer used is momentum SGD. For normal distribution estimators, the sampling proportion $\gamma$ for PatchCore is set to 0.1, and the low-rank constant $D'$ for OrthoAD is set to 100, by default.

\begin{table*}[t]
\centering
\caption{Comparison of \texttt{CAReg} with its corresponding baseline 
normal distribution estimation methods on MPDD with $K=\{2,4,8\}$. AUC in \% averaged over 10 runs for each category and the standard deviation (SD) of PC and PC + \texttt{CAReg} in percentage (\% is omitted) over these 10 runs (10 different support sets) are provided, together with a macro-average score over all categories. For each pair, the best-performing method is marked in bold.}
\label{tal:mpdd}
\vspace{-5pt}
\scriptsize
\scalebox{0.95}{
\setlength{\tabcolsep}{0.6pt}{
\begin{tabular}{C{1.75cm}C{0.9cm}>{\columncolor{gray!15}}C{1.06cm}C{0.85cm}>{\columncolor{gray!15}}C{0.88cm}C{0.85cm}>{\columncolor{gray!15}}C{0.88cm}C{0.9cm}>{\columncolor{gray!15}}C{1.06cm}C{0.85cm}>{\columncolor{gray!15}}C{0.88cm}C{0.85cm}>{\columncolor{gray!15}}C{0.88cm}C{0.9cm}>{\columncolor{gray!15}}C{1.06cm}C{0.85cm}>{\columncolor{gray!15}}C{0.88cm}C{0.85cm}>{\columncolor{gray!15}}C{0.88cm}}
\toprule
\multirow{3}{*}{Category} & \multicolumn{6}{c}{K=2} & \multicolumn{6}{c}{K=4} & \multicolumn{6}{c}{K=8}\\
\cmidrule(lr){2-7} \cmidrule(lr){8-13} \cmidrule(lr){14-19}
& \makecell[c]{PaDim\\\cite{defard2021padim}} & \makecell[c]{PaDim+\\CAReg} & \makecell[c]{OAD\\\cite{orthoad}} & \makecell[c]{OAD+\\CAReg} & \makecell[c]{PC\\\cite{patchcore}} & \makecell[c]{PC+\\CAReg} & \makecell[c]{PaDim\\\cite{defard2021padim}} & \makecell[c]{PaDim+\\CAReg} & \makecell[c]{OAD\\\cite{orthoad}} & \makecell[c]{OAD+\\CAReg} & \makecell[c]{PC\\\cite{patchcore}} & \makecell[c]{PC+\\CAReg} & \makecell[c]{PaDim\\\cite{defard2021padim}} & \makecell[c]{PaDim+\\CAReg} & \makecell[c]{OAD\\\cite{orthoad}} & \makecell[c]{OAD+\\CAReg} & \makecell[c]{PC\\\cite{patchcore}} & \makecell[c]{PC+\\CAReg}\\
\cmidrule(lr){1-19}
bracket black     & 45.3 & \textbf{60.5} & 52.8 & \textbf{57.9} & 47.6 \tiny 11.0 & \textbf{52.7} \tiny \textbf{10.6} & 52.5 & \textbf{62.8} & 50.3 & \textbf{56.9} & 53.7 \tiny \textbf{7.5} & \textbf{58.3} \tiny 9.3  & 50.5 & \textbf{62.6} & 54.5 & \textbf{65.1} & 53.1 \tiny \textbf{5.0} & \textbf{60.7} \tiny 6.8 \\
bracket brown      & 43.5 & \textbf{56.6} & 41.4 & \textbf{60.3} & 56.6 \tiny 3.6  & \textbf{57.2} \tiny \textbf{3.4} & 48.8 & \textbf{62.9} & 54.0 & \textbf{60.4} & 62.9 \tiny 5.6 & \textbf{67.6} \tiny \textbf{5.3} & 48.8 & \textbf{64.5} & 61.9 & \textbf{63.3} & 64.6 \tiny \textbf{3.3} & \textbf{72.3} \tiny 5.0 \\
bracket white    & \textbf{59.2} & 59.0 & 35.9 & \textbf{46.0} & 43.6 \tiny 16.6 & \textbf{53.2} \tiny \textbf{7.0} & \textbf{63.3} & 61.0 & 44.7 & \textbf{45.6} & \textbf{53.3} \tiny 15.9 & 52.8 \tiny \textbf{7.6} & 55.9 & \textbf{75.1} & 49.0 & \textbf{67.3} & \textbf{63.0} \tiny 6.0 & 62.6 \tiny \textbf{5.8} \\
connector     & 42.3 & \textbf{83.6} & \textbf{57.4} & 35.7 & 66.5 \tiny 12.2 & \textbf{80.8} \tiny \textbf{1.0} & 50.1 & \textbf{75.0} & 50.2 & \textbf{51.2} & 72.1 \tiny 13.3 & \textbf{87.5} \tiny \textbf{5.6} & 76.2 & \textbf{80.1} & 51.0 & \textbf{65.2} & 84.9 \tiny 4.8 & \textbf{95.7} \tiny \textbf{0.6} \\
metal plate       & 35.3 & \textbf{56.2} & 43.6 & \textbf{93.6} & 89.6 \tiny 6.5 & \textbf{96.1} \tiny \textbf{0.4} & 38.7 & \textbf{79.5} & 42.7 & \textbf{95.6} & 95.4 \tiny 5.9 & \textbf{99.4} \tiny \textbf{0.3} & 38.9 & \textbf{87.7} & 38.2 & \textbf{95.3} & 98.8 \tiny 1.7 & \textbf{99.9} \tiny \textbf{0.0} \\
tubes   & 72.2 & \textbf{73.1} & \textbf{60.5} & 53.3 & 47.0 \tiny 10.6 & \textbf{63.9} \tiny \textbf{4.4} & \textbf{68.8} & 65.7 & \textbf{59.7} & 52.3 & 49.5 \tiny 6.4 & \textbf{68.9} \tiny \textbf{3.9} & 62.5 & \textbf{64.9} & 63.2 & \textbf{64.1} & 54.8 \tiny 4.6 & \textbf{72.5} \tiny \textbf{4.0} \\
\cmidrule(lr){1-19}
Average    & 49.6 & \textbf{64.8} & 48.6 & \textbf{57.8} & 58.5 \tiny 10.1 & \textbf{67.3} \tiny \textbf{4.8} & 53.7 & \textbf{67.8} & 50.3 & \textbf{60.3} & 64.5 \tiny 9.1 & \textbf{72.4} \tiny \textbf{5.3} & 55.5 & \textbf{72.5} & 53.0 & \textbf{70.1} & 69.9 \tiny 4.2 & \textbf{77.3} \tiny \textbf{3.7} \\
\bottomrule
\end{tabular}
}}
\end{table*}

\begin{table*}[t]
\centering
\caption{Comparison of \texttt{CAReg} with state-of-the-art FSAD methods and vanilla AD methods (upper-bound using the entire dataset for training) on MVTec and MPDD. The macro-average of AUC (in \%) over all categories in each dataset is reported. For FSAD, results of different shots are reported ($K=\{2,4,8\}$).}
\label{tal:vanilla}
\vspace{-5pt}
\small
\scalebox{0.98}{
\setlength{\tabcolsep}{1.0pt}{
\begin{tabular}{C{1.5cm}C{1.8cm}C{2.0cm}C{2.0cm}C{1.8cm}C{2.0cm}C{1.4cm}C{1.4cm}C{1.4cm}C{1.4cm}C{1.4cm}}
\toprule
& \multicolumn{1}{c}{\multirow{2}{*}{Data}} 
& \multicolumn{1}{c}{\multirow{2}{*}{Method}}
& \multicolumn{1}{c}{\multirow{2}{*}{Year}} & \multicolumn{1}{c}{\multirow{2}{*}{Pretrain}} & \multicolumn{1}{c}{\multirow{2}{*}{Backbone}} & \multicolumn{2}{c}{MVTec} & \multicolumn{2}{c}{MPDD} \\
& & & & & & image & pixel & image & pixel\\
\cmidrule(lr){1-10}
& \multicolumn{1}{l}{\multirow{5}{*}{$\quad$K=2}} & \multicolumn{1}{l}{$\quad$RegAD~\cite{regad}} & 2022 & ImageNet & Res18 & 85.7 & 94.6 & 63.4 & 93.2\\
& & \multicolumn{1}{l}{$\quad$RFR~\cite{lee2023few}} & 2023 & ImageNet & Res18 & 86.6 & 95.9 & - & -\\
& & \multicolumn{1}{l}{$\quad$PACKD~\cite{gu2023prototype}} & 2023 & ImageNet & WRN50 & 90.2 & 95.0 & 66.6 & 94.4\\
& & \multicolumn{1}{l}{$\quad$PromptAD~\cite{li2024promptad}} & 2024 & CLIP & ViTB/16 & \textbf{91.2} & - & - & -\\
& & \multicolumn{1}{l}{$\quad$CAReg (ours)} & 2023 & ImageNet & Res18 & 90.2 & \textbf{96.2} & \textbf{67.3} & \textbf{94.9}\\
\cmidrule(lr){2-10}

& \multicolumn{1}{l}{\multirow{5}{*}{$\quad$K=4}} & \multicolumn{1}{l}{$\quad$RegAD~\cite{regad}} & 2022 & ImageNet & Res18 & 88.2 & 95.8 & 68.3 & 93.9\\
Few-Shot & & \multicolumn{1}{l}{$\quad$RFR~\cite{lee2023few}} & 2023 & ImageNet & Res18 & 89.3 & 96.4 & - & -\\
AD & & \multicolumn{1}{l}{$\quad$PACKD~\cite{gu2023prototype}} & 2023 & ImageNet & WRN50 & 91.6 & 96.2 & 69.8 & \textbf{94.8}\\
& & \multicolumn{1}{l}{$\quad$PromptAD~\cite{li2024promptad}} & 2024 & CLIP & ViTB/16 & 92.7 & - & - & -\\
& & \multicolumn{1}{l}{$\quad$CAReg (ours)} & 2023 & ImageNet & Res18 & \textbf{93.7} & \textbf{97.0} & \textbf{72.4} & 94.4\\
\cmidrule(lr){2-10}

& \multicolumn{1}{l}{\multirow{5}{*}{$\quad$K=8}} & \multicolumn{1}{l}{$\quad$RegAD~\cite{regad}} & 2022 & ImageNet & Res18 & 91.2 & 96.8 & 71.9 & 95.1\\
& & \multicolumn{1}{l}{$\quad$RFR~\cite{lee2023few}} & 2023 & ImageNet & Res18 & 91.9 & 96.9 & - & -\\
& & \multicolumn{1}{l}{$\quad$PACKD~\cite{gu2023prototype}} & 2023 & ImageNet & WRN50 & \textbf{95.3} & 97.3 & 70.5 & 95.3\\
& & \multicolumn{1}{l}{$\quad$PromptAD~\cite{li2024promptad}} & 2024 & CLIP & ViTB/16 & 93.1 & - & - & -\\
& & \multicolumn{1}{l}{$\quad$CAReg (ours)} & 2023 & ImageNet & Res18 & 95.1 & \textbf{97.4} & \textbf{77.3} & \textbf{95.8}\\
\cmidrule(lr){1-10}

& \multicolumn{1}{l}{$\quad$full data} & \multicolumn{1}{l}{$\quad$GANomaly~\cite{ganomaly}} & 2018  & / & UNet & 80.5 & - & 64.8 & -\\
& \multicolumn{1}{l}{$\quad$full data} & \multicolumn{1}{l}{$\quad$ARNet~\cite{ARNet}} & 2022  & / & UNet & 83.9 & - & 69.7 & -\\
& \multicolumn{1}{l}{$\quad$full data} & \multicolumn{1}{l}{$\quad$MKD~\cite{MKD}} & 2021  & ImageNet & Res18 & 87.7 & 90.7 & - & - \\
Vanilla & \multicolumn{1}{l}{$\quad$full data} & \multicolumn{1}{l}{$\quad$CutPaste~\cite{cutpaste}} & 2021  & ImageNet & Res18 & 95.2 & 96.0 & - & -\\
AD & \multicolumn{1}{l}{$\quad$full data} & \multicolumn{1}{l}{$\quad$FYD~\cite{focus}} & 2022  & ImageNet & Res18 & 97.3 & 97.4 & - & -\\
& \multicolumn{1}{l}{$\quad$full data} & \multicolumn{1}{l}{$\quad$PaDiM~\cite{defard2021padim}} & 2021  & ImageNet & WRN50 & 97.9 & 97.5 & 74.8 & 96.7\\
& \multicolumn{1}{l}{$\quad$full data} & \multicolumn{1}{l}{$\quad$PatchCore~\cite{patchcore}} & 2022  & ImageNet & WRN50 & 99.1 & 98.1 & 82.1 & 95.7\\
& \multicolumn{1}{l}{$\quad$full data} & \multicolumn{1}{l}{$\quad$CflowAD~\cite{cflow}} & 2022  & ImageNet & WRN50 & 98.3 & 98.6 & 86.1 & 97.7\\
\bottomrule
\end{tabular}
}}
\end{table*}

\vspace{-2mm}
\subsection{Comparison with State-of-the-art FSAD Methods}
Table~\ref{tal:sum} presents the comparison results on MVTec and MPDD datasets. The results show that the two extended methods, TDG+ and DiffNet+, achieve only limited improvements (TDG+: $\leq3.8\%$, DiffNet+: $\leq2.1\%$) compared to their corresponding original versions trained without data from multiple categories. However, the proposed category-agnostic registration training method (\texttt{CAReg}) demonstrates significant improvements compared to its three baseline methods, PaDim~\cite{defard2021padim}, OrthoAD~\cite{orthoad}, and Patchcore~\cite{patchcore}. In the few-shot scenarios with $K={2,4,8}$, \texttt{CAReg} outperforms PaDim by 10.8\%, 11.2\%, 10.7\% on MVTec, and 15.2\%, 14.1\%, 17.0\% on MPDD, respectively. Moreover, compared to the current state-of-the-art method Patchcore, \texttt{CAReg} improves the average AUC by 4.4\%, 4.9\%, 3.8\% on MVTec, and 8.8\%, 7.9\%, 7.4\% on MPDD. Individual category-wise results are shown in Table~\ref{tal:mvtec} for MVTec and in Table~\ref{tal:mpdd} for MPDD.

\begin{table*}[t]
\centering
\caption{Ablation study of \texttt{CAReg} on MVTec and MPDD. Factors under analysis are: data augmentation on support set (DA);  feature registration aggregated training (FR); the spatial transformer networks (STN); and accumulated feature (AF). PaDim is used as the normal distribution estimator. The macro-average AUC (in \%) over all categories and over 10 runs is reported, with the best-performing setting for each shot marked in bold.}
\label{tal:abl_all}
\small
\scalebox{0.98}{
\setlength{\tabcolsep}{1.5pt}{
\begin{tabular}{C{1.0cm}C{1.0cm}C{1.0cm}C{1.03cm}|C{1.03cm}C{1.03cm}C{1.03cm}C{1.03cm}C{1.03cm}C{1.03cm}|C{1.03cm}C{1.03cm}C{1.03cm}C{1.03cm}C{1.03cm}C{1.03cm}}
\toprule
\multicolumn{4}{c|}{\multirow{2}{*}{Modules}}
& \multicolumn{6}{c|}{MVTec}                                        & \multicolumn{6}{c}{MPDD}                                          \\
\cmidrule(lr){5-16}
& & & & \multicolumn{3}{c}{image} & \multicolumn{3}{c|}{pixel} & \multicolumn{3}{c}{image} & \multicolumn{3}{c}{pixel}  \\
\cmidrule(lr){1-16}
 DA & FR & STN & AF & K=2 & K=4 & K=8               & K=2 & K=4 & K=8                  & K=2 & K=4 & K=8               & K=2 & K=4 & K=8                   \\
\cmidrule(lr){1-16}
            &            &            &            & 74.7 & 78.0 & 80.5 & 88.6 & 90.5 & 92.1 & 49.6 & 53.7 & 55.5 & 89.5 & 91.2 & 92.0 \\
 \checkmark &            &            &            & 81.5 & 84.9 & 87.4 & 93.3 & 94.7 & 95.5 & 50.8 & 54.2 & 61.1 & 92.4 & 93.3 & 93.9\\
            & \checkmark &            &            & 78.0 & 80.9 & 83.1 & 90.8 & 92.5 & 94.0 & 53.9 & 55.5 & 57.2 & 91.5 & 92.2 & 93.0 \\
            & \checkmark & \checkmark &            & 79.1 & 82.9 & 84.9 & 90.5 & 93.3 & 94.3 & 57.6 & 60.9 & 62.7 & 91.0 & 91.8 & 93.0\\
 \checkmark & \checkmark &            &            & 83.0 & 86.4 & 89.3 & 94.7 & 95.9 & 96.6 & 52.8 & 57.7 & 64.8 & 93.3 & 94.1 & 94.4\\
 \checkmark & \checkmark & \checkmark &            & \textbf{85.7} & 88.2 & \textbf{91.2} & 94.6 & 95.8 & 96.7 & 63.4 & \textbf{68.8} & 71.9 & 93.2 & 93.9 & 95.1 \\
 \checkmark & \checkmark & \checkmark & \checkmark & 85.5 & \textbf{89.2} & \textbf{91.2} & \textbf{95.6} & \textbf{96.2} & \textbf{97.1} & \textbf{64.8} & 67.8 & \textbf{72.5} & \textbf{93.7} & \textbf{94.6} & \textbf{95.2} \\
\bottomrule
\end{tabular}}}
\end{table*}

Without any parameter fine-tuning, \texttt{CAReg} is tested after category-agnostic registration training. In contrast, other methods that use separate models for each category only focus on optimizing the performance for individual categories in each experiment. As a result, it can be challenging to achieve the best result for every category with \texttt{CAReg}. However, when averaging the results for all categories and all shot scenarios, \texttt{CAReg} consistently outperforms the corresponding baselines of PaDim, OAD, and PC by 10.9\%, 8.5\%, and 4.4\% on MVTec. \texttt{CAReg} also shows the lowest standard deviation in performance across the 15 categories of MVTec. For instance, when $K=8$, \texttt{CAReg} achieves standard deviations of 10.54\%, 9.52\%, and 6.34\% on the three estimators (PaDim, OrthoAD, and PatchCore), which are significantly lower than their corresponding baseline methods, indicating improved generalizability of the proposed method across different categories. Also, \texttt{CAReg} achieves an impressive AUC of 95.1\% on the MVTec dataset with $K=8$, representing a remarkable $\approx$7\% improvement compared to Metaformer~\cite{metaformer}. While Metaformer has its advantages due to the use of an additional large-scale dataset (MSRA10K~\cite{msra10k}) during training, \texttt{CAReg} demonstrates its effectiveness without relying on such extra data.

For real-world applications of FSAD, the time required to adapt a pre-trained model to a target category (referred to as adaptation time) is an important factor. In the case of TDG+ and DiffNet+, the fine-tuning procedure is very time-consuming as it involves updating parameters for many epochs. With \texttt{CAReg}, adaptation speed is significantly faster. This is because statistical estimators like PaDim, OrthoAD, and PatchCore can be used directly after inference for the support images, without the need for parameter fine-tuning. As a result, \texttt{CAReg} achieves the fastest adaptation speed among the compared methods, making it more practical and efficient for real-world applications. When averaging the adaptation time on MVTec and MPDD for FSAD with $K={2,4,8}$, PaDim+CAReg achieves the fastest adaptation speed (4.47s) compared to DiffNet+ (357.75s) and TDG+ (1559.76s).

To explain the impact of the support set’s distribution on model performance, we report the standard deviation (SD) in percentage of PC and PC + \texttt{CAReg} for each category across 10 support sets, as shown in Table~\ref{tal:mvtec} and Table~\ref{tal:mpdd} (smaller font). Our proposed method, \texttt{CAReg}, significantly reduces the performance variance caused by different support sets. This stabilization is especially evident in the MPDD dataset, where varying support sets have a more pronounced impact compared to the MVTec dataset, which has a naturally lower baseline SD. By averaging results across multiple sets, we enhance the reliability and robustness of our findings, thereby strengthening the credibility of our results.

\vspace{-2mm}
\subsection{Comparison with State-of-the-art Vanilla AD Methods}
While vanilla AD methods have an inherent advantage as they use the entire training dataset and train separate models for each category, \texttt{CAReg} with the PaDim estimator still achieves competitive performance, as shown in Table~\ref{tal:vanilla}. The state-of-the-art vanilla AD methods GANomaly~\cite{ganomaly}, ARNet~\cite{ARNet}, MKD~\cite{MKD}, CutPaste~\cite{cutpaste}, FYD~\cite{focus}, PaDiM~\cite{defard2021padim}, PatchCore~\cite{patchcore} and CflowAD~\cite{cflow} are considered. Even with only 4 support images from MVTec, \texttt{CAReg} (89.2\% AUC) outperforms MKD (87.7\%) using the same ResNet-18 backbone. As the number of support images increases to 128, \texttt{CAReg} achieves an impressive AUC of 95.9\%. Similarly, on the MPDD dataset with 128 support images, \texttt{CAReg} (83.2\% AUC) surpasses PatchCore (82.1\%), which uses a deeper backbone (WRN50).

\subsection{Ablation Studies}\label{sec:abl}
The ablation study was conducted to investigate the importance of four key components used in \texttt{CAReg}: (1) data augmentations applied to the support sets (DA), (2) feature registration aggregated training on multiple categories (FR), (3) the spatial transformer networks (STN), and (4) the accumulated feature registration (AF). These modules were combined based on a systematic experimental process. We performed extensive ablation studies to assess the impact of each component individually and collectively.

\subsubsection{Feature Registration Aggregated Training} 
Table~\ref{tal:abl_all} demonstrates that the Feature Registration (FR) module yields improvements in AUC of 3.3\%, 2.9\%, and 2.6\% for $K={2,4,8}$ on MVTec, and 4.3\%, 1.8\%, and 1.7\% on MPDD, respectively. These findings indicate that feature registration aggregated training on multiple categories enhances the model's ability to generalize to novel categories. Moreover, the effectiveness of the registration aggregated training (FR) is evident, regardless of the presence or absence of data augmentation (DA). 

\subsubsection{Spatial Transformation Modules}
The utilization of Spatial Transformer Networks (STNs) enables the application of large-scale transformations, leading to improved feature registration. For instance, the STN module contributes to an AUC improvement from 89.3\% to 91.2\% on MVTec and from 64.8\% to 71.9\% on MPDD ($K=8$), as demonstrated in Table~\ref{tal:abl_all}. Additionally, the incorporation of STN modules shows comparable performance in terms of pixel-level localization. However, it is important to note that when STN is used, an inverse transformation operation is necessary to realign the transformed features with their original images, which may result in some pixel-level imprecision.

\begin{table*}[t]
\centering
\caption{Ablation study of augmentation selection on MVTec ($K=2$). For each selected category, \emph{i.e.}, three objects (capsule, pill, and toothbrush) and three textures (carpet, leather, and tile), AUC (in \%) averaged over 10 runs is reported, with best-performing settings marked in \textbf{bold} and second-best-performing settings \underline{underlined}.}
\label{tal:abl_augsel}
\small
\setlength{\tabcolsep}{2.0pt}{
\begin{tabular}{C{1.5cm}C{2.2cm}|C{4.0cm}|C{2.9cm}|C{1.85cm}C{1.85cm}C{1.85cm}}
\toprule
& \multirow{2}{*}{Category} & Augmentations & \multirow{2}{*}{All Augmentations} & \multicolumn{3}{c}{Selected Augmentations}\\
&  & Aligned with RegAD~\cite{regad} &  & KL & JS & Wasserstein\\
\cmidrule(lr){1-7} 
& Capsule & 68.6 & \underline{68.9} & \underline{68.9} & 68.8 &\textbf{69.4} \\
Object & Pill & 66.1 & \underline{73.3} & 71.7 & \textbf{73.4} & \underline{73.3} \\
& Toothbrush & 86.8 & 87.1 & \textbf{89.3} & 87.8 & \textbf{89.3} \\
\cmidrule(lr){1-7}
& Carpet & 96.7 & 96.4 & \underline{96.9} & 96.5 & \textbf{97.0} \\
Texture & Leather & \textbf{100} & \textbf{100} & 99.7 & 99.7 & \textbf{100}\\
& Tile & \textbf{98.9} & 98.2 & 97.4 & 97.6 & \underline{98.5} \\
\bottomrule
\end{tabular}
\vspace{-8pt}
}
\end{table*}

\begin{table}[t]
\centering
\caption{Comparison of \texttt{CAReg} with or without augmentations in $K=2$. False positive rates (FPR) in \% over 10 runs with a macro-average score over all categories are reported.}
\begin{tabular}{ccc}
\toprule
False Positive Rates  & MVTec  & MPDD   \\ \cmidrule(lr){1-3}
w/o augmentation & 5.4\% & 29.3\% \\
w/ augmentation  & \textbf{2.7\%} & \textbf{19.8\%} \\
\bottomrule
\end{tabular}
\label{tal:FPR}
\end{table}

\begin{table*}[t]
\centering
\caption{Evaluation of cross-dataset model generalization for \texttt{CAReg} on MVTec and MPDD with $K=\{2,4,8\}$. The terms `orig.' and `cross' indicate whether the model was trained on the original or a different dataset: the actual training dataset is indicated in brackets. The AUC (in \%) averaged over 10 runs is reported.}
\label{tal:cross}
\small
\scalebox{0.97}{
\setlength{\tabcolsep}{1.5pt}{
\begin{tabular}{C{1.3cm}C{1.0cm}|C{1.35cm}C{1.35cm}C{1.35cm}C{0.75cm}|C{1.35cm}C{1.35cm}C{1.35cm}C{0.75cm}|C{1.35cm}C{1.35cm}C{1.35cm}C{0.75cm}}
\toprule
Test on & Shots &
PaDim & \multicolumn{3}{c|}{PaDim+CAReg} & OAD & \multicolumn{3}{c|}{OAD+CAReg} & PC & \multicolumn{3}{c}{PC+CAReg} \\
\cmidrule(lr){1-14} 
 & & & orig. (MVTec) & cross (MPDD) & $\triangle$  & & orig. (MVTec) & cross (MPDD) & $\triangle$ & & orig. (MVTec) & cross (MPDD) & $\triangle$\\
 \cmidrule(lr){3-14}
 \multirow{3}{*}{MVTec} & K=2 & 74.7 & 85.5 & 83.3 & -2.2 & 76.7 & 83.6 & 82.8 & -0.8 & 85.8 & 90.2 & 89.4 & -0.8\\
& K=4     & 78.0 & 89.2 & 87.2 & -2.0 & 77.6 & 86.9 & 86.5 & -0.4 & 88.8 & 93.7 & 92.9 & -0.8\\
& K=8     & 80.5 & 91.2 & 89.4 & -1.8 & 79.8 & 89.2 & 88.9 & -0.3 & 91.3 & 95.1 & 94.3 & -0.8\\
\cmidrule(lr){1-14}
 & & & orig. (MPDD) & cross (MVTec) & $\triangle$  & & orig. (MPDD) & cross (MVTec) & $\triangle$ & & orig. (MPDD) & cross (MVTec) & $\triangle$\\
  \cmidrule(lr){3-14}
 \multirow{3}{*}{MPDD} & K=2 & 49.6 & 64.8 & 63.0 & -1.8 & 48.6 & 57.8 & 57.3 & -0.5 & 58.5 & 67.3 & 74.4 & +7.1\\
& K=4     & 53.7 & 67.8 & 70.9 & +3.1 & 50.3 & 60.3 & 65.1 & +4.8 & 64.5 & 72.4 & 77.7 & +5.3\\
& K=8     & 55.5 & 72.5 & 73.1 & +0.6 & 53.0 & 70.1 & 68.0 & -2.1 & 69.9 & 77.3 & 80.5 & +3.2\\
\bottomrule
\end{tabular}}}
\end{table*}

\begin{table*}[t]
\centering
\caption{Anomaly \textbf{detection/localization} results for cross-dataset model generalization on the MVTec and MPDD datasets with $K=\{2,4,8\}$ using  \texttt{CAReg} with the PC normal distribution estimator. The term `orig.' indicates that the model was trained on the original dataset, while `cross' indicates that the model was trained on the other dataset. AUC in \% averaged over 10 runs for each category is reported, together with a macro-average score over all categories. The best-performing result is marked in bold, and the second-best result is underlined.}
\label{tal:cross_detail_pc}
\scriptsize
\scalebox{0.97}{
\setlength{\tabcolsep}{1.5pt}{
\begin{tabular}{C{1.8cm}|C{1.25cm}C{1.25cm}>{\columncolor{gray!15}}C{1.25cm}C{1.35cm}|C{1.25cm}C{1.25cm}>{\columncolor{gray!15}}C{1.25cm}C{1.35cm}|C{1.25cm}C{1.25cm}>{\columncolor{gray!15}}C{1.25cm}C{1.35cm}}
\toprule
\multirow{4}{*}{Category} & \multicolumn{4}{c}{K=2} & \multicolumn{4}{c}{K=4} & \multicolumn{4}{c}{K=8}\\
\cmidrule(lr){2-5} \cmidrule(lr){6-9} \cmidrule(lr){10-13}
& \multirow{2}{*}{PC} & \multicolumn{3}{c}{PC+CAReg} & \multirow{2}{*}{PC} & \multicolumn{3}{c}{PC+CAReg} & \multirow{2}{*}{PC} & \multicolumn{3}{c}{PC+CAReg}\\
\cmidrule(lr){3-5} \cmidrule(lr){7-9} \cmidrule(lr){11-13}
&& orig. & cross & $\triangle$ && orig. & cross & $\triangle$ && orig. & cross & $\triangle$\\
\cmidrule(lr){1-13}
\multicolumn{13}{c}{\textbf{\textsl{Test on MVTec Dataset}}} \\
\cmidrule(lr){1-13}
Bottle     & 99.8/98.1 & 99.6/98.7 & 99.9/98.2 & +0.3/-0.5 & 99.4/98.1 & 99.9/98.7 & 100/98.5 & +0.1/-0.2 & 100/98.1 & 99.9/98.9 & 100/98.5 & +0.1/-0.4 \\
Cable      & 86.2/95.7 & 91.5/95.9 & 91.4/96.0 & -0.1/+0.1 & 88.9/97.0 & 93.8/96.6 & 94.1/96.6 & +0.3/+0.0 & 88.8/97.2 & 95.2/96.8 & 95.3/96.8 & +0.1/+0.0 \\
Capsule    & 66.2/97.3 & 70.5/97.0 & 69.7/96.1 & -0.8/-0.9 & 76.1/98.0 & 81.4/97.9 & 80.9/97.8 & -0.5/-0.1 & 88.7/98.5 & 82.0/98.0 & 83.4/98.2 & +1.4/+0.2 \\
Carpet     & 99.0/98.9 & 99.4/98.8 & 97.9/98.7 & -1.5/-0.1 & 99.1/98.9 & 99.0/98.8 & 98.5/98.7 & -0.5/-0.1 & 99.1/98.9 & 99.3/98.9 & 98.5/98.7 & -0.8/-0.2 \\
Grid       & 58.3/65.5 & 79.9/83.7 & 74.5/80.6 & -5.4/-3.1 & 62.2/65.1 & 89.8/89.2 & 85.7/88.0 & -4.1/-1.2 & 70.1/74.3 & 91.0/90.5 & 86.0/89.2 & -5.0/-1.3 \\
Hazelnut   & 89.5/95.6 & 97.9/98.6 & 98.4/98.0 & +0.5/-0.6 & 98.8/96.4 & 99.4/98.6 & 98.6/98.7 & -0.8/+0.1 & 99.5/97.7 & 99.8/98.7 & 99.4/98.7 & -0.4/+0.0 \\
Leather    & 100/99.1 & 100/99.1 & 99.5/98.8 & -0.5/-0.3 & 100/99.0 & 100/99.1 & 99.7/98.9 & -0.3/-0.2 & 100/99.1 & 100/99.1 & 99.9/98.8 & -0.1/-0.3 \\
Metal Nut  & 92.0/95.5 & 97.7/98.2 & 94.7/97.4 & -3.0/-0.8 & 92.4/97.3 & 98.2/98.9 & 96.9/98.0 & -1.3/-0.9 & 95.1/97.9 & 98.1/98.9 & 96.7/98.3 & -1.4/-0.6 \\
Pill       & 83.7/96.2 & 86.1/97.9 & 83.6/95.6 & -2.5/-2.3 & 84.3/96.6 & 90.8/98.1 & 86.7/96.2 & -4.1/-1.9 & 89.3/97.0 & 90.4/98.6 & 86.2/96.3 & -4.2/-2.3 \\
Screw      & 45.2/87.9 & 62.2/94.6 & 61.8/94.6 & -0.4/+0.0 & 51.1/88.9 & 73.7/96.3 & 72.9/96.3 & -0.8/+0.0 & 47.6/91.9 & 80.9/97.5 & 79.7/97.5 & -1.2/+0.0 \\
Tile       & 98.9/94.4 & 100/96.6 & 99.9/95.2 & -0.1/-1.4 & 99.3/95.1 & 100/96.7 & 99.9/95.3 & -0.1/-1.4 & 99.2/95.1 & 99.9/96.6 & 99.9/95.1 & +0.0/-1.5 \\
Toothbrush & 86.4/97.3 & 83.0/97.9 & 85.0/97.6 & +2.0/-0.3 & 87.2/97.3 & 86.6/98.3 & 91.0/98.2 & +4.4/-0.1 & 95.8/98.3 & 94.8/98.7 & 96.6/98.6 & +1.8/-0.1  \\
Transistor & 85.0/94.0 & 91.2/93.7 & 89.3/92.5 & -1.9/-1.2 & 97.1/94.9 & 97.8/95.3 & 94.5/94.7 & -3.3/-0.6 & 100/95.5 & 99.2/96.8 & 98.3/96.0 & -0.9/-0.8 \\
Wood       & 98.9/92.7 & 99.0/94.6 & 98.6/91.6 & -0.4/-3.0 & 98.0/92.5 & 98.6/94.6 & 97.9/92.1 & -0.7/-2.5 & 97.8/92.9 & 98.6/94.9 & 98.4/92.1 & -0.2/-2.8 \\
Zipper     & 98.3/97.7 & 95.2/97.5 & 96.2/98.1 & +1.0/+0.6 & 98.3/97.9 & 96.5/97.6 & 96.6/97.6 & +0.1/+0.0 & 98.0/98.0 & 96.8/97.7 & 96.9/97.7 & +0.1/+0.0 \\
\cmidrule(lr){1-13}
Average    & 85.8/93.7 & \textbf{90.2}/\textbf{96.2} & \underline{89.4}/\underline{95.3} & -0.8/-0.9 & 88.8/94.2 & \textbf{93.7}/\textbf{97.0} & \underline{92.9}/\underline{96.4} & -0.8/-0.6 & 91.3/95.4 & \textbf{95.1}/\textbf{97.4} & \underline{94.3}/\underline{96.7} & -0.8/-0.7 \\
\cmidrule(lr){1-13}
\multicolumn{13}{c}{\textbf{\textsl{Test on MPDD Dataset}}} \\
\cmidrule(lr){1-13}
bracket black & 47.6/89.0 & 52.7/91.5 & 50.8/90.2 & -1.9/-1.3 & 53.7/90.5 & 58.3/92.5 & 56.8/92.2 & -1.5/-0.3 & 53.1/91.1 & 60.7/92.6 & 56.1/91.4 & -4.6/-1.2  \\
bracket brown & 56.6/92.2 & 57.2/94.7 & 57.8/94.7 & +0.6/+0.0 & 62.9/94.6 & 67.6/96.3 & 67.6/96.3 & +0.0/+0.0 & 64.6/96.0 & 72.3/92.1 & 72.6/97.2 & +0.3/+5.1 \\
bracket white       & 43.6/92.3 & 53.2/92.1 & 63.6/94.7 & +10.4/+2.6 & 53.3/93.7 & 52.8/94.5 & 65.7/97.2 & +12.9/+2.7 & 63.0/95.2 & 62.6/96.1 & 71.2/98.0 & +8.6/+1.9 \\
connector     & 66.5/95.9 & 80.8/98.4 & 98.0/94.0 & +17.2/-4.4 & 72.1/97.0 & 87.5/98.7 & 97.1/95.7 & +9.6/-3.0 & 84.9/98.1 & 95.7/99.2 & 99.5/96.1 & +3.8/-3.1 \\
metal plate     & 89.6/95.8 & 96.1/97.3 & 99.4/98.2 & +3.3/+0.9 & 95.4/96.8 & 99.4/98.0 & 99.8/98.5 & +0.4/+0.5 & 98.8/97.8 & 99.9/98.2 & 100/98.5 & +0.1/+0.3 \\
tubes     & 47.0/92.7 & 63.9/95.3 & 76.9/95.7 & +13.0/+0.4 & 49.5/93.8 & 68.9/95.9 & 79.3/96.2 & +10.4/+0.3 & 54.8/94.6 & 72.5/96.4 & 83.1/97.1 & +10.6/+0.7 \\
\cmidrule(lr){1-13}
Average    & 58.5/93.0 & \underline{67.3}/\textbf{94.9} & \textbf{74.4}/\underline{94.6} & +7.1/-0.3 & 64.5/\underline{94.4} & \underline{72.4}/\underline{94.4} & \textbf{77.7}/\textbf{96.0} & +5.3/+1.6 & 69.9/95.5 & \underline{77.3}/\underline{95.8} & \textbf{80.5}/\textbf{96.4} & +3.2/+0.6 \\
\bottomrule
\end{tabular}}
}
\end{table*}

To investigate the impact of the degree of freedom in the transformation, we conducted experiments with different STN versions, and the results are presented in the appendix. Notably, for MVTec, STNs that involve rotation+scale transformations exhibit the best performance. This observation aligns with the fact that most objects in this dataset are centrally located, resulting in similar spatial positions. Consequently, the inclusion of translation in the STNs does not significantly contribute to performance improvement, and the overall effect of the STN modules is limited. In contrast, for MPDD, STNs lead to more substantial performance gains, particularly since large-scale transformations are often required. The images in this dataset exhibit various spatial orientations and positions for objects, making it beneficial to align the features with affine transformations. As a result, the STNs that allow affine transformations achieve the best performance in this scenario.

\subsubsection{Accumulated Feature Registration}
Unlike RegAD~\cite{regad}, which performs registration between two individual images, \texttt{CAReg} introduces an innovative approach that registers one image to a set of images, thereby enhancing the robustness of the registration process. As shown in Table~\ref{tal:abl_all}, accumulated feature (AF) registration leads to improvements in anomaly localization AUC by 1.0\%, 0.4\%, and 0.4\% on MVTec, and 0.5\%, 0.7\%, and 0.1\% on MPDD, for $K=2,4,8$, respectively, confirming the effectiveness of the approach.

\subsubsection{Data Augmentations and Selection Methods}\label{sec:abl_aug} 
Applying augmentations to the support set during testing is crucial for improved performance. As shown in Table~\ref{tal:abl_all}, with $K=2,4,8$, the AUC of AD improves by 1.2\%, 0.5\%, and 0.6\% on MPDD, and by 6.8\%, 6.9\%, and 6.9\% on MVTec, respectively. Default augmentations like rotation, translation, flipping, and graying are used, following practices in~\cite{TDG} and~\cite{regad}. However, augmentations such as cutpaste and mixup are excluded due to their potential to simulate anomalies~\cite{cutpaste}.

\begin{figure}[t]
\centering
\includegraphics[width=0.4\textwidth]{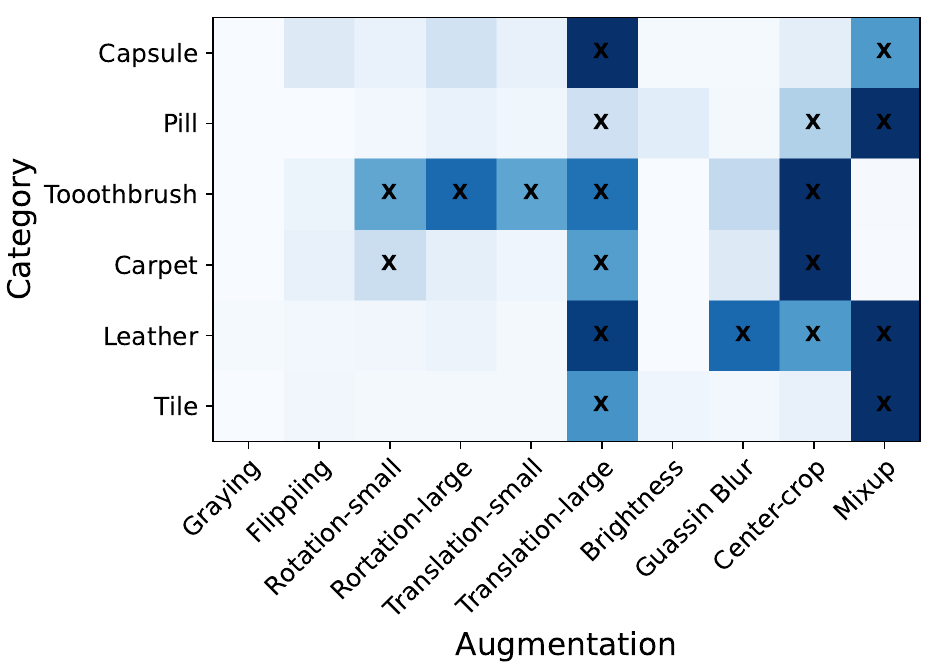}
\vspace{-6pt}
\caption{Visualization of Wasserstein distances and augmentation selection results. Darker color corresponds to a larger Wasserstein distance. Augmentations marked with $\times$ were removed from augmentations for that category.}
\vspace{-10pt}
\label{img:aug}
\end{figure}
 
To explore suitable augmentations for each category, we experimented with 10 augmentations: graying, flipping, large/small-angle rotation, large/small-scale translation, brightness, Gaussian blur, center-crop, and mixup. Three selection methods were used: (i) RegAD-aligned augmentations~\cite{regad}, (ii) all augmentations, and (iii) our proposed Wasserstein distance-based selection. Experiments were conducted on six categories, including three objects (capsule, pill, toothbrush) and three textures (carpet, leather, tile). 

\begin{figure*}[t]
\centering
\includegraphics[width=0.9\textwidth]{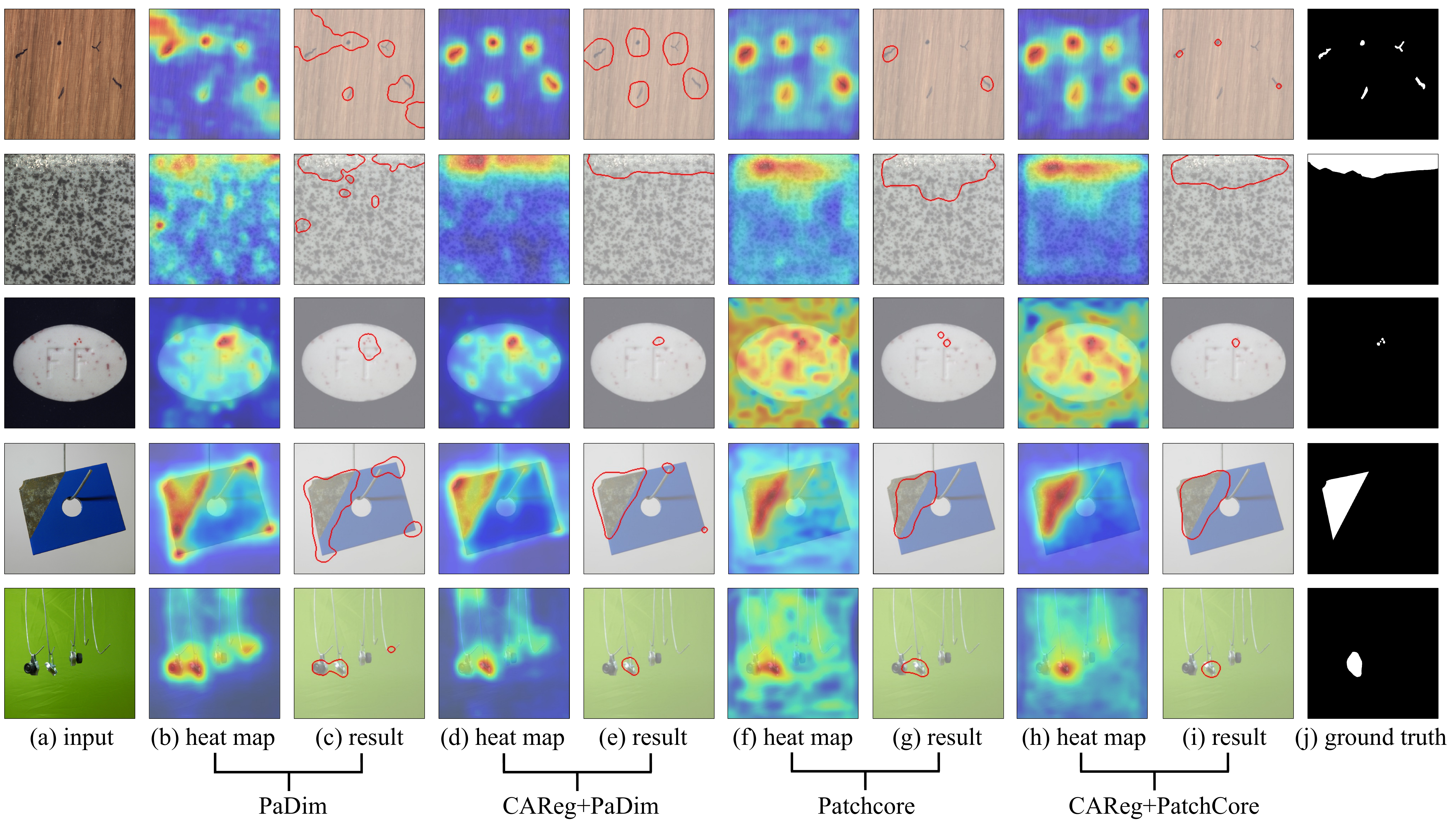}
\caption{Visualization of anomaly localization results on MVTec (top three rows) and MPDD (bottom two rows). (a) Input. PaDim (b) heat map and (c) result. CAReg+PaDim (d) heat map and (e) result. Patchcore (f) heat map and (g) result. CAReg+PatchCore (h) heat map and (i) result. (j) GT.}
\label{img:result}
\end{figure*}

\begin{figure*}[t]
\begin{minipage}[t]{0.5\textwidth}
\centering\includegraphics[width=8.7cm]{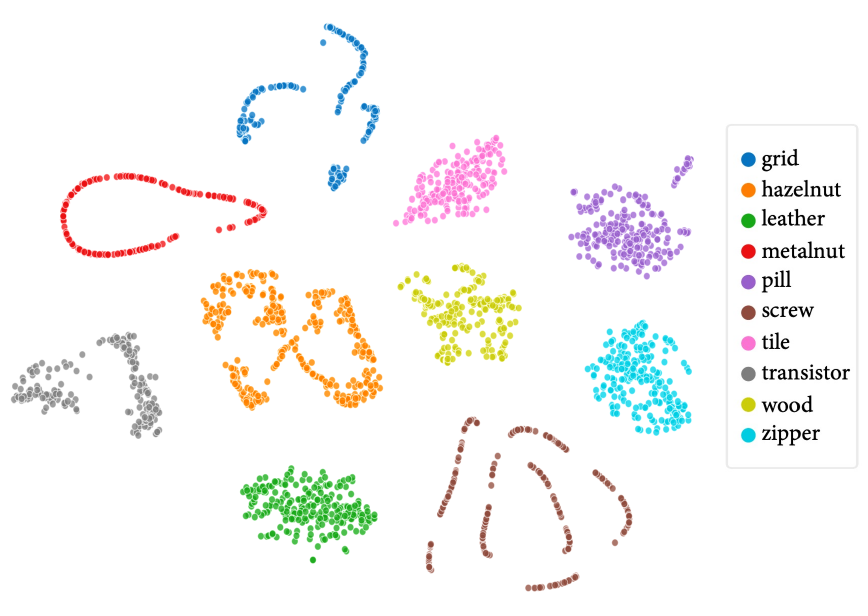}

(a) Without Feature Registration
\end{minipage}
\begin{minipage}[t]{0.5\textwidth}
\centering
\includegraphics[width=7.0cm]{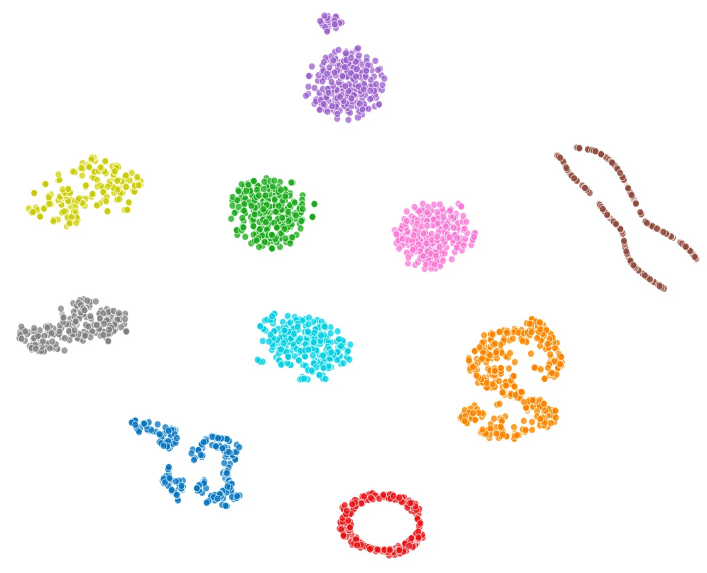}

(b) \texttt{CAReg}
\end{minipage}
\caption{The t-SNE visualization of features learned from MVTec. The same t-SNE optimization iterations are used for both methods. \texttt{CAReg} leads to more compacted feature distributions for each category, and larger separation among categories. (a) Without feature registration. \texttt{CAReg}.}
\label{img:tsne}
\end{figure*}

Fig.~\ref{img:aug} displays the selection results for each category obtained Fig.~\ref{img:aug} shows the selection results for each category using the Wasserstein distance-based method. Table~\ref{tal:abl_augsel} indicates that our method, despite using fewer augmentations, outperforms using all augmentations, highlighting the importance of selecting suitable augmentations. Data augmentations have varying effects on objects and textures. For the three objects, performance significantly improves with appropriate augmentations, whereas the selection has little impact on the three textures. This smaller impact on textures may be due to the difficulty in altering their characteristic distribution with simple augmentations. Remarkably, using only four fixed augmentations (graying, flipping, small-angle rotation, small-scale translation) achieves comparable results, especially for tiles, where it performs best. 

To further evaluate augmentation selection, we included results using KL and JS distances, alongside the Wasserstein distance. The comparison in Table~\ref{tal:abl_augsel} shows that KL and JS distances generally perform better than RegAD’s augmentations and similarly or slightly worse than the Wasserstein distance method. This suggests that KL and JS distances could be promising alternatives, offering competitive performance.

In addition, we also addressed concerns about possible increased false positive rates from non-independent augmented examples. As shown in Table~\ref{tal:FPR}, thorough evaluations show that our augmentation strategy interestingly reduces the false positive rate (FPR) from 5.44\% to 2.65\% on MVTec and from 29.3\% to 19.8\% on MPDD on average with $K=2$ (detailed in the appendix). This reduction is largely due to our augmentation selection module, designed to enrich the dataset while avoiding biases or artifacts that could degrade image quality. By carefully selecting beneficial augmentation methods, we maintain dataset integrity and enhance overall model performance.

\subsection{Cross-dataset Generalization}

In real-world applications FSAD methods need to be capable of performing accurately even when the test images differ significantly from those used during training. To evaluate the ability of \texttt{CAReg} to generalise in this way we performed experiments where the network was trained on one dataset and tested on another. For these cross-dataset generalization experiments we used the MVTec and MPDD datasets as these differ significantly in terms of categories, backgrounds, and variability (as shown in Table~\ref{tal:datasetsum}). The results are shown in Table~\ref{tal:cross}. These results effectively demonstrate that altering the training dataset has only a marginal impact on performance when compared to training on the original datasets. For instance, when assessing the performance on MVTec using the PC estimator, transitioning from the original training data to MPDD (a notably smaller dataset in terms of categories and samples compared to MVTec) leads to a mere 0.8\% reduction in AUC. Conversely, when evaluating on MPDD, training the model on MVTec yields performance enhancements. An underlying reason for this improved performance could be attributed to MVTec's broader range of categories, which equips \texttt{CAReg} with a more diverse array of classes to learn from. This augmented diversity enables \texttt{CAReg} to better grasp the overarching concepts of category-agnostic registration, consequently enhancing its efficacy in handling previously unseen categories.

The detailed results of AD/localization for each category can be found in Table~\ref{tal:cross_detail_pc}, offering insights into AD and localization outcomes under the PC estimator. It serves to validate the robustness of registration learning. This is evident from the fact that even when the distribution of the training data is altered, a significant improvement compared to the baseline is sustained. For instance, when evaluating on MVTec with K=2, CAReg demonstrates a performance enhancement of 4.4\%/2.5\% when trained on its original dataset (MVTec). Even when transitioning to a distinct distribution (MPDD) for training, the improvement remains substantial at 3.6\%/1.6\%. In essence, the results firmly establish the \textit{proposed method's ability to generalize across datasets}, showcasing how the influence of training with an entirely different dataset is minimal due to the inherent category-agnostic nature of the registration training approach.

\subsection{Visualization Analysis}

Fig.~\ref{img:result} provides qualitative analysis results to showcase the effectiveness of the category-agnostic feature registration training approach and its impact on AD and localization performance. With the PaDim normal distribution estimator, the anomaly localization produced by \texttt{CAReg} (column e) is observed to be closer to the ground truth (GT) (column j) compared to its corresponding baseline (column c). Similarly, the results produced by \texttt{CAReg} with the PatchCore normal distribution estimator (column i) are closer to GT than those produced by its corresponding baseline (column g). These visualizations illustrate the effectiveness of the training procedure via feature registration aggregated on multiple categories, resulting in improved anomaly localization.

In Fig.~\ref{img:tsne}, t-SNE~\cite{maaten2008visualizing} is used to visualize the features extracted on the MVTec test set, with each dot representing a normal image. It is evident that after the feature registration training, the features become more compact within each category. Simultaneously, features from different categories are better separated. The presence of more compact feature distributions is desirable, as it enhances the accuracy of the normal distribution estimate for each category. This improved separation of features contributes to better discrimination between normal and anomalous samples, leading to enhanced AD performance. Overall, the visualization results provide qualitative evidence of the efficacy of the category-agnostic feature registration training approach in enhancing AD and localization. 
\section{Conclusion}
\label{sec:conclusion}

This paper introduces a novel training method, called \texttt{CAReg}, for FSAD, by learning generalizable registration techniques across different categories using only normal images for each category. This allows the model to accurately register a test image with its corresponding support (normal) images from unseen novel categories without the need for re-training or parameter fine-tuning. The experimental results demonstrate that \texttt{CAReg} outperforms state-of-the-art FSAD methods in both AD and anomaly localization tasks on standard benchmark datasets. Even when compared to AD methods trained with much larger volumes of data, \texttt{CAReg} remains competitive. It significantly enhances both the accuracy and efficiency of FSAD, showcasing its high potential for real-world AD applications.


\bibliographystyle{IEEEtran}
\bibliography{egbib}

\begin{IEEEbiography}[{\includegraphics[width=1in,height=1.25in,clip,keepaspectratio]{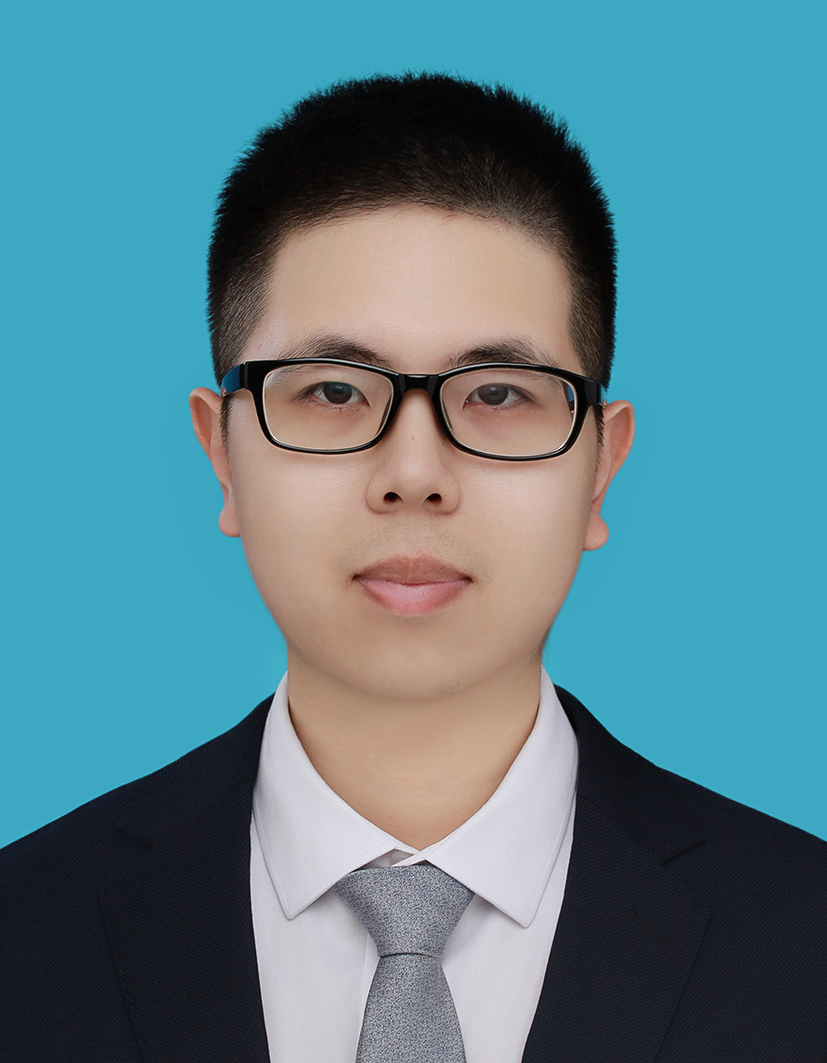}}]{Chaoqin Huang} received the B.Eng. degree in computer science from Shanghai Jiao Tong University (SJTU), Shanghai, China, in 2019, and the Ph.D. degree in information and communication engineering, through a joint Ph.D. degree program, from SJTU and the National University of Singapore (NUS), Singapore, in 2024.

He is currently an Assistant Professor with the College of Information Communications, National University of Defense Technology (NUDT), Wuhan, China. His research interests include computer vision and anomaly detection.
\end{IEEEbiography}
\vspace{-8pt}
\begin{IEEEbiography}[{\includegraphics[width=1in,height=1.25in,clip,keepaspectratio]{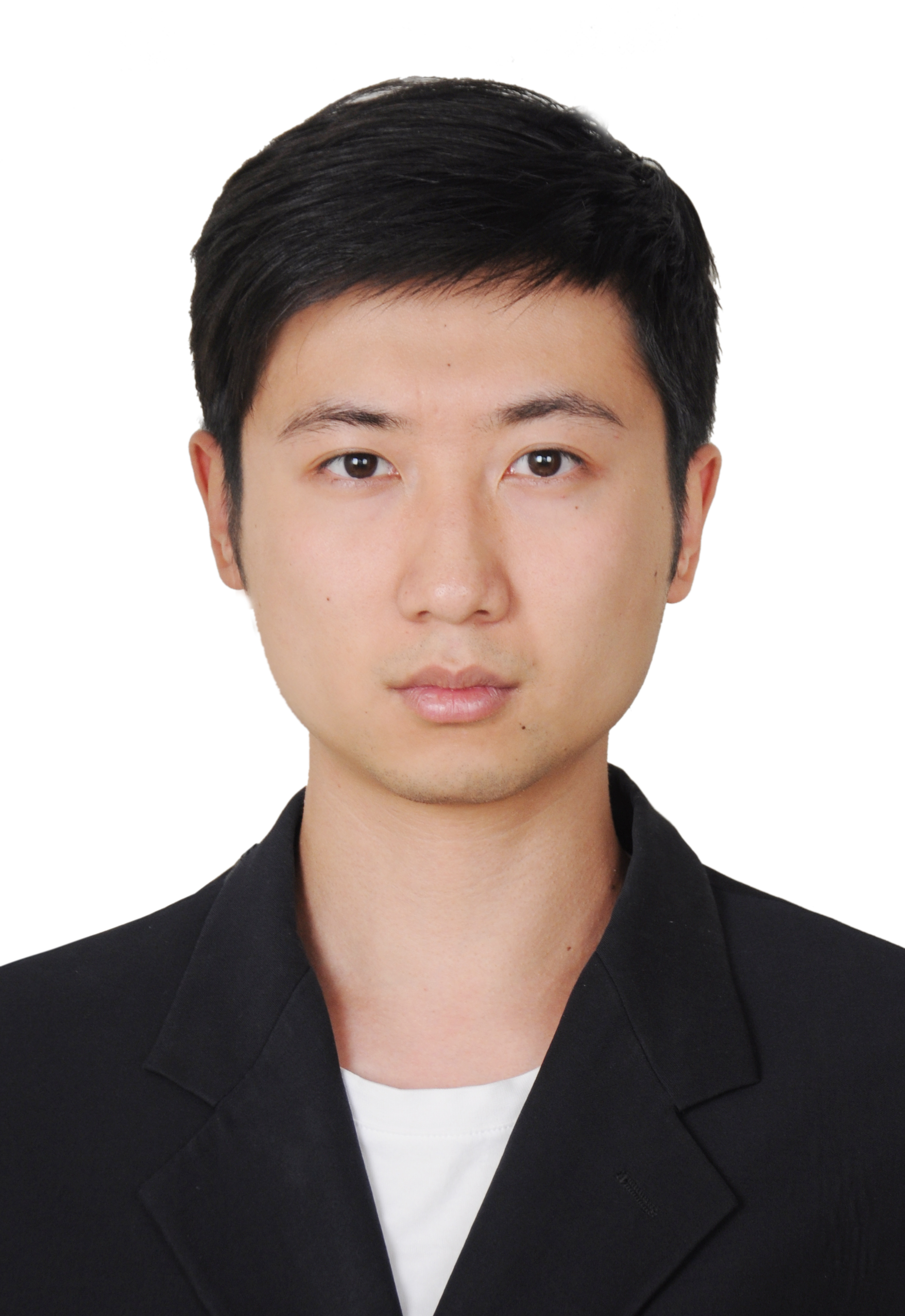}}]{Haoyan Guan} received the B.Eng. degree from Harbin Institute of Technology, Harbin, China, the M.Sc. degree from Shanghai Jiao Tong University, Shanghai, China, and the Ph.D. degree in computer science from the Department of Informatics, King’s College London, London, U.K.

His research is concerned with the limited supervision in computer vision.
\end{IEEEbiography}
\vspace{-8pt}
\begin{IEEEbiography}[{\includegraphics[width=1in,height=1.25in,clip,keepaspectratio]{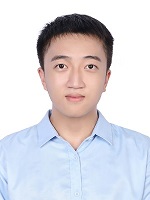}}]{Aofan Jiang} received the B.Eng. degree from Shanghai Jiao Tong University (SJTU), Shanghai, China, in 2022, where he is currently pursuing the M.Sc. degree in information engineering with the Department of Electronic Engineering.

His research is concerned with anomaly detection in computer vision.
\end{IEEEbiography}

\begin{IEEEbiography}[{\includegraphics[width=1in,height=1.25in,clip,keepaspectratio]{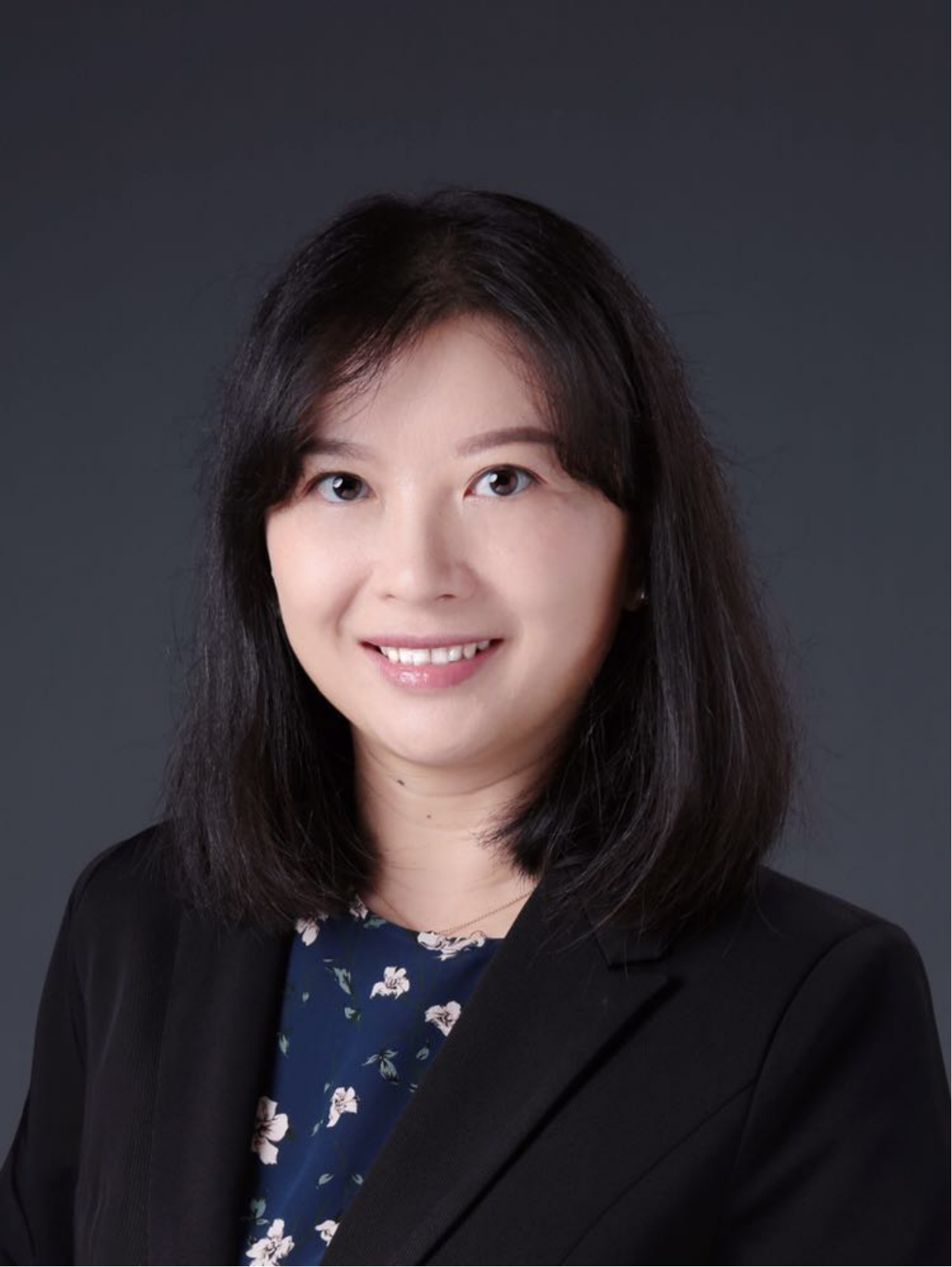}}]{Ya Zhang} received the bachelor’s degree from Tsinghua University, Beijing, China, and the Ph.D. degree in information sciences and technology from Pennsylvania State University, University Park, PA,
USA.

Her career has spanned academia and industry, including roles as a Tenure-Track Assistant Professor at The University of Kansas, Lawrence, KS, USA, and a Senior Research and Development Manager at Yahoo! Labs, where she led initiatives to improve search quality. She currently holds key positions as the Vice Dean of the Digital Medical Research Institute and the Deputy Director of the Intelligent Healthcare Research Center, Shanghai Artificial Intelligence Laboratory, Shanghai, China. She is also a Distinguished Professor with the School of Artificial Intelligence, Shanghai Jiao Tong University, Shanghai. She has authored more than 200 scholarly articles in high-impact international journals and conferences, with her work being cited more than 10 000 times. Her research focuses on machine learning applications in multimedia and healthcare.

Dr. Zhang’s achievements are recognized through various awards and honors, including named a Leading Talent under the National High-Level Talent Special Support Program in 2023, receiving the First Prize of the Shanghai Technological Invention Award in 2022 and the EURASIP Best Paper Award in 2019, and recognized as an Outstanding Doctoral Dissertation Advisor by the Chinese Association for Artificial Intelligence in 2018.
\end{IEEEbiography}

\begin{IEEEbiography}[{\includegraphics[width=1in,height=1.25in,keepaspectratio]{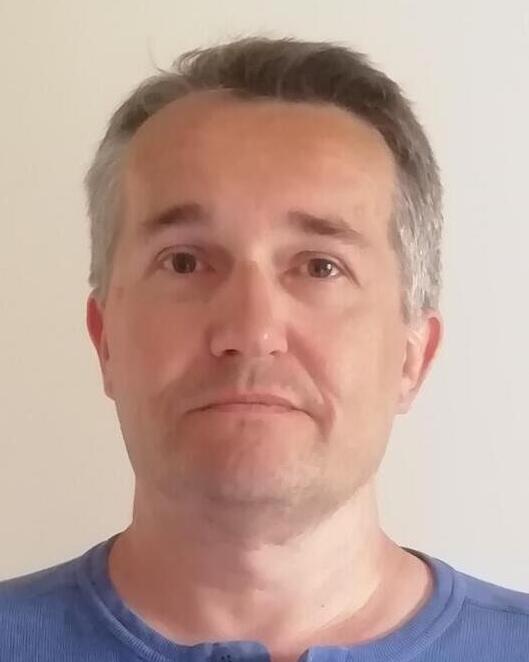}}]{Michael Spratling} received the B.Eng. degree in engineering science from Loughborough University, Loughborough, U.K., and the M.Sc. and Ph.D. degrees in artificial intelligence and neural computation from The University of Edinburgh, Edinburgh, U.K.

He is currently a Reader in computational neuroscience and visual cognition with the Department of Informatics, King’s College London, London, U.K. His research is concerned with understanding the computational and neural mechanisms underlying visual perception and developing biologically inspired neural networks to solve problems in computer vision and machine learning.

\end{IEEEbiography}
\vspace{-8pt}
\begin{IEEEbiography}[{\includegraphics[width=1in,height=1.25in,keepaspectratio]{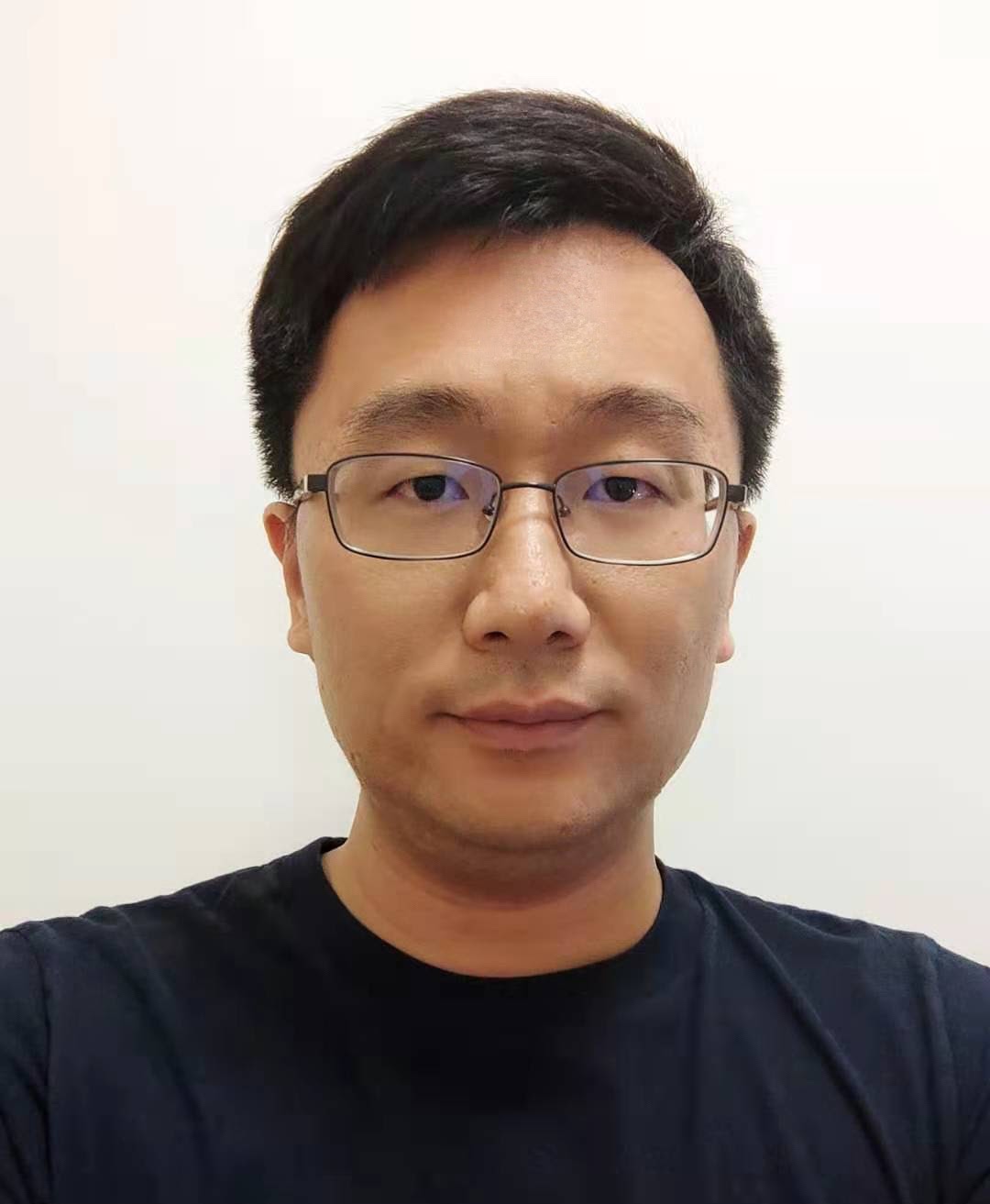}}]{Xinchao Wang} (Senior Member, IEEE) received the Ph.D. degree from the Ecole Polytechnique Federale de Lausanne (EPFL), Lausanne, Switzerland, and the first-class Honorable degree from The Hong Kong Polytechnic University (HKPU), Hong Kong.

He was a Tenure-Track Assistant Professor of computer science with the Stevens Institute of Technology, Hoboken, NJ, USA, and a Swiss-NSF Post-Doctoral Fellow with the University of Illinois Urbana–Champaign (UIUC), Champaign, IL, USA, with Prof. Thomas S. Huang. He is currently an Assistant Professor with the Department of Electrical and Computer Engineering (ECE), National University of Singapore (NUS), Singapore. His articles have been published in major venues including CVPR, ICCV, ECCV, NeurIPS, ICLR, AAAI, IJCAI, ACL, MICCAI, TPAMI, IJCV, TIP, TKDE, and TMI. His research interests include artificial intelligence, computer vision, machine learning, medical image analysis, and multimedia.

Dr. Wang regularly serves as an Area Chair for CVPR, ICCV, ECCV,
NeurIPS, ICML, ICPR, ICIP, and ICME, and as a Senior Program Committee Member of AAAI and IJCAI. He has been or is currently serving as an Associate Editor for IEEE TIP, IEEE TCSVT, JVCI, and PR.
\end{IEEEbiography}

\begin{IEEEbiography}[{\includegraphics[width=1in,height=1.25in,clip,keepaspectratio]{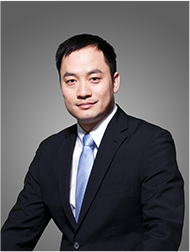}}]
{Yanfeng Wang} (Member, IEEE) received the B.S. degree from PLA Information Engineering University, Zhengzhou, China, and the master’s and Ph.D. degrees in business management from Shanghai Jiao Tong University, Shanghai, China.

He is currently the Deputy Dean of the School of Artificial Intelligence, Shanghai Jiao Tong University. He has long been focused on scientific research and innovation at the intersection of artificial intelligence with media and healthcare, as well as the translation of research outcomes.

Dr. Wang’s achievements have been recognized with the First Prize of the Shanghai Science and Technology Progress Award (twice), the First Prize of the Shanghai Technological Invention Award (once), and the First Prize of the China Institute of Electronics Science and Technology Award (once). He has also been honored as a 2022 Shanghai Outstanding Academic Leader and a recipient of the 2015 Shanghai May Fourth Youth Medal.
\end{IEEEbiography}


\clearpage
\newpage
\onecolumn
\appendix

\begin{algorithm*}[th]
\small
\caption{Anomaly score measurement with PatchCore}\label{alg:aug}
    \SetKwInOut{KwIn}{Input}
    \SetKwInOut{KwOut}{Output}
    \KwIn{CAReg pre-trained $\phi$, few-shot support set $\mathcal{S}$, test sample $I_{test}$, data augmentation operator $\alpha$, patch feature extractor $\mathcal{P}$, memory size target $l$, random linear projection $\psi$.}
    \KwOut{The anomaly score map $d$ for the test sample.}
    $\mathcal{M} \leftarrow \left\{ \right\}$ \textcolor{gray}{//Memory bank initialization}\\
    \For{$x_{i} \in \mathcal{S}$}
    {
        $\mathcal{M} \leftarrow \mathcal{M} \cup \mathcal{P}(\phi(x_{i})) \cup \mathcal{P}(\phi(\alpha(x_{i})))$\;
    }
    $\mathcal{M}_{C} \leftarrow \left\{ \right\}$ \textcolor{gray}{//Apply coreset sampling for memory bank}\\
    \For{$i \in \left [0,\cdots, l-1  \right ]$}
    {
        $m_{i} \leftarrow \underset{m \in \mathcal{M} - \mathcal{M_{C}}}{\argmax} \underset{n \in \mathcal{M_{C}}}{\min} \left\| \psi(m) - \psi(n) \right\|_{2}$\;
        $\mathcal{M_{C}} \leftarrow \mathcal{M_{C}} \cup \left\{m_{i} \right\}$\;
    }
    $\mathcal{M} \leftarrow \mathcal{M_{C}}$\;
    $\left\{ f_{ij} \right\}_{(i,j)\in [1,W]\times [1,H]} \leftarrow \mathcal{P}(\phi(I_{test}))$ \textcolor{gray}{//Registerd patch features}\\
    \For{each position $(i,j)$}
    {
        $m^* \leftarrow \argmin_{m \in \mathcal{M}}\left\|f_{ij}-m\right\|_2$ \textcolor{gray}{//Nearest neighbor search}\\
        $d_{ij} \leftarrow \left\|f_{ij}-m^*\right\|_2$\;
    }
    $\mathit{d} \leftarrow (\mathit{d}_{ij})_{1\leqslant i\leqslant W, 1\leqslant j\leqslant H}$. 
\end{algorithm*}

\begin{table*}[h]
\centering
\caption{Ablation study with different STN modules on MVTec and MPDD ($K=2$). The macro-average AUC (in \%) over all categories and over 10 runs is reported, with the best-performing setting marked in bold.}
\label{tal:abl_stn}
\small
\scalebox{0.98}{
\setlength{\tabcolsep}{1.0pt}{
\begin{tabular}{C{1.5cm}C{1.4cm}C{1.8cm}C{1.4cm}C{1.2cm}C{1.2cm}C{1.4cm}C{1.8cm}C{1.8cm}C{2.3cm}C{1.4cm}}
\toprule
Data & w/o STN & translation & rotation & scale & shear & \makecell[c]{rotation\\+scale} & \makecell[c]{translation\\+scale} & \makecell[c]{translation\\+rotation} & \makecell[c]{translation\\+rotation+scale} & affine \\
\cmidrule(lr){1-11}
MVTec & 83.0 & 84.5 & 85.0 & 84.9 & 84.9 & \textbf{85.7} & 84.9 & 84.2 & 84.9 & 84.5 \\
MPDD & 52.8 & 62.3 & 57.7 & 59.2 & 59.0 & 61.5 & 61.8 & 61.0 & 61.7 & \textbf{63.4} \\
\bottomrule
\end{tabular}}}
\end{table*}

\begin{table*}[h]
\centering
\caption{Comparison of \texttt{CAReg} (w/ AUG) with its no augmentation (w/o AUG) on MVTec and MPDD with $K=2$. False positive rates (FPR) in \% over 10 runs for each category is reported. For each pair, the best-performing method is marked in bold.}
\label{APXtal:FPR}
\small
\scalebox{0.98}{
\setlength{\tabcolsep}{1.0pt}{
\begin{tabular}{C{1.8cm}C{2.2cm}C{2.2cm}C{2.2cm}C{1.8cm}C{1.8cm}C{1.8cm}C{1.8cm}C{1.8cm}}
\toprule
Method & Bottle & Cable & Capsule & Carpet & Grid & Hazelnut & Leather & Metal Nut \\
\cmidrule(lr){1-9}
w/o AUG & \textbf{0.00\%} & 1.04\% & 10.90\% & \textbf{0.00\%} & 22.50\% & 4.15\% & \textbf{0.00\%} & 7.94\%  \\
w/ AUG  & \textbf{0.00\%} & \textbf{0.00\%} & \textbf{2.94\%} & \textbf{0.00\%} & \textbf{8.45\%} & \textbf{0.00\%} & \textbf{0.00\%} & \textbf{0.15\%}  \\ 
\cmidrule(lr){1-9}
 & Pill & Screw & Tile & Toothbrush & Transistor & Wood & Zipper & \textbf{Average} \\
\cmidrule(lr){1-9}
w/o AUG & \textbf{0.00\%} & \textbf{26.8\%} & \textbf{0.00\%} & 7.23\% & 1.01\% & \textbf{0.00\%} & \textbf{0.00\%} & 5.44\%  \\
w/ AUG            & \textbf{0.00\%} & 27.5\% & \textbf{0.00\%} & \textbf{0.67\%} & \textbf{0.00\%} & \textbf{0.00\%} & \textbf{0.00\%} & \textbf{2.65\%}   \\ 
\bottomrule
 & Bracket Black & Bracket Brown & Bracket White & Connector & Metal Plate & Tubes & \textbf{Average} &  \\
\cmidrule(lr){1-9}
w/o AUG  & 43.1\% & 32.2\% & 42.1\% & 42.9\% & 5.3\% & 10.1\% & 29.3\% &   \\
w/ AUG            & \textbf{42.7\%} & \textbf{31.8\%} & \textbf{38.3\%} & \textbf{5.9\%} & \textbf{0.3\%} & \textbf{0.0\%} & \textbf{19.8\%} &  \\ 
\bottomrule
\end{tabular}}}
\end{table*}

\vfill

\end{document}